\documentclass[conference]{IEEEtran}

\usepackage{silence}
\usepackage[utf8]{inputenc}
\usepackage[T1]{fontenc}

\usepackage{graphicx}
\usepackage{placeins}  % prevents moving float objects (\FloatBarrier)

\usepackage{subfigure}
\usepackage[table]{xcolor}

\usepackage{amssymb,amsmath,amsthm}
\usepackage{booktabs}
\usepackage{algorithm}
\usepackage{algorithmic}
\usepackage[switch]{lineno}

\usepackage{array}
\usepackage{booktabs}
\usepackage{ragged2e}
\usepackage{makecell}
\usepackage{adjustbox}
\usepackage{threeparttable}
\usepackage{multicol,multirow}

\usepackage[hidelinks]{hyperref}
\usepackage{url}
\begin{document}

\title{Masked strategies for images with small objects}

\author{\IEEEauthorblockN{Anonymous Authors}}

\author{
\IEEEauthorblockN{
H. Martin Gillis$^{1}$,
Ming Hill$^{2}$, 
Paul Hollensen$^{3}$, 
Alan Fine$^{3,4,5}$, and
Thomas Trappenberg$^{1*}$
}
\vspace{-0.2cm} \\
\IEEEauthorblockA{
$^1$Faculty of Computer Science, Dalhousie University, Halifax, NS, Canada
}
\IEEEauthorblockA{
$^2$Department of Computer Science, Boston University, Boston, MA, United States
}
\IEEEauthorblockA{
$^3$Alentic Microscience Inc., Halifax, NS, Canada
}
\IEEEauthorblockA{
$^4$Department of Physiology and Biophysics, Dalhousie University, Halifax, NS, Canada
}
\IEEEauthorblockA{
$^5$School of Biomedical Engineering, Dalhousie University, Halifax, NS, Canada 
\vspace{0.2cm}
}
\{martin.gillis, isaac.xu, a.fine\}@dal.ca,
minghill@bu.edu,
phollensen@alentic.com,
tt@cs.dal.ca
}
\maketitle

\renewcommand{\thefootnote}{\fnsymbol{footnote}}
\footnotetext[1]{Corresponding author:~\href{mailto:tt@cs.dal.ca}{tt@cs.dal.ca}}
\renewcommand{\thefootnote}{\arabic{footnote}}

% \linenumbers

% ABSTRACT
\begin{abstract}

The hematology analytics used for detection and classification of small blood components is a significant challenge.  
In particular, when objects exists as small pixel-sized entities in a large context of similar objects.
Deep learning approaches using supervised models with pre-trained weights (\textit{e.g.}, ImageNet), such as residual networks (ResNets) and vision transformers (ViTs) have demonstrated success for many applications.
Unfortunately, when applied to images outside the domain of learned representations, these methods often result with less than acceptable performance.
A strategy to overcome this can be achieved by using self-supervised models, where representations are learned from in-domain images and weights are then applied for downstream applications, such as semantic segmentation.
Recently, masked autoencoders (MAEs) have proven to be effective in pre-training ViTs to obtain representations that captures global context information.
By masking regions of an image and having the model learn to reconstruct both the masked and non-masked regions, the resulting weights can be used for various applications.
However, if the sizes of the objects in images are less than the size of the mask (\textit{i.e.}, patch size), the global context information is lost, making it almost impossible to reconstruct the image.
In this study, we investigated the effect of mask ratios  and patch sizes for blood components using a ``small-scale'' MAE to obtain learned ViT encoder representations.
We then applied the encoder weights to train a U-Net Transformer (UNETR) for semantic segmentation to obtain both local and global contextual information.
Our experimental results demonstrates that both smaller mask ratios and patch sizes improve the reconstruction of images using a MAE.
We also show the results of semantic segmentation with and without pre-trained weights, where smaller-sized blood components benefited with pre-training. 
Overall, our proposed method offers an efficient and effective strategy for the segmentation and classification of small objects. 

\end{abstract}

% INTRODUCTION
\section{Introduction}
\label{sect:introduction}
The identification and quantification of blood components plays an important role in evaluating both personal and public health~\cite{WHO::2021a,Gurcan::2009a}.
This information provides healthcare professionals the knowledge required make diagnoses and raise awareness for potential personal, local, and/or global health related concerns.
Healthcare professionals typically obtain a blood sample from a patient and then send the sample offsite for analysis, which can take several hours or days to process.
These delays can present increased risks for time-sensitive scenarios and currently there are limited options available.

\begin{figure}[!htb]
    \centering
    \includegraphics[width=0.42\textwidth]{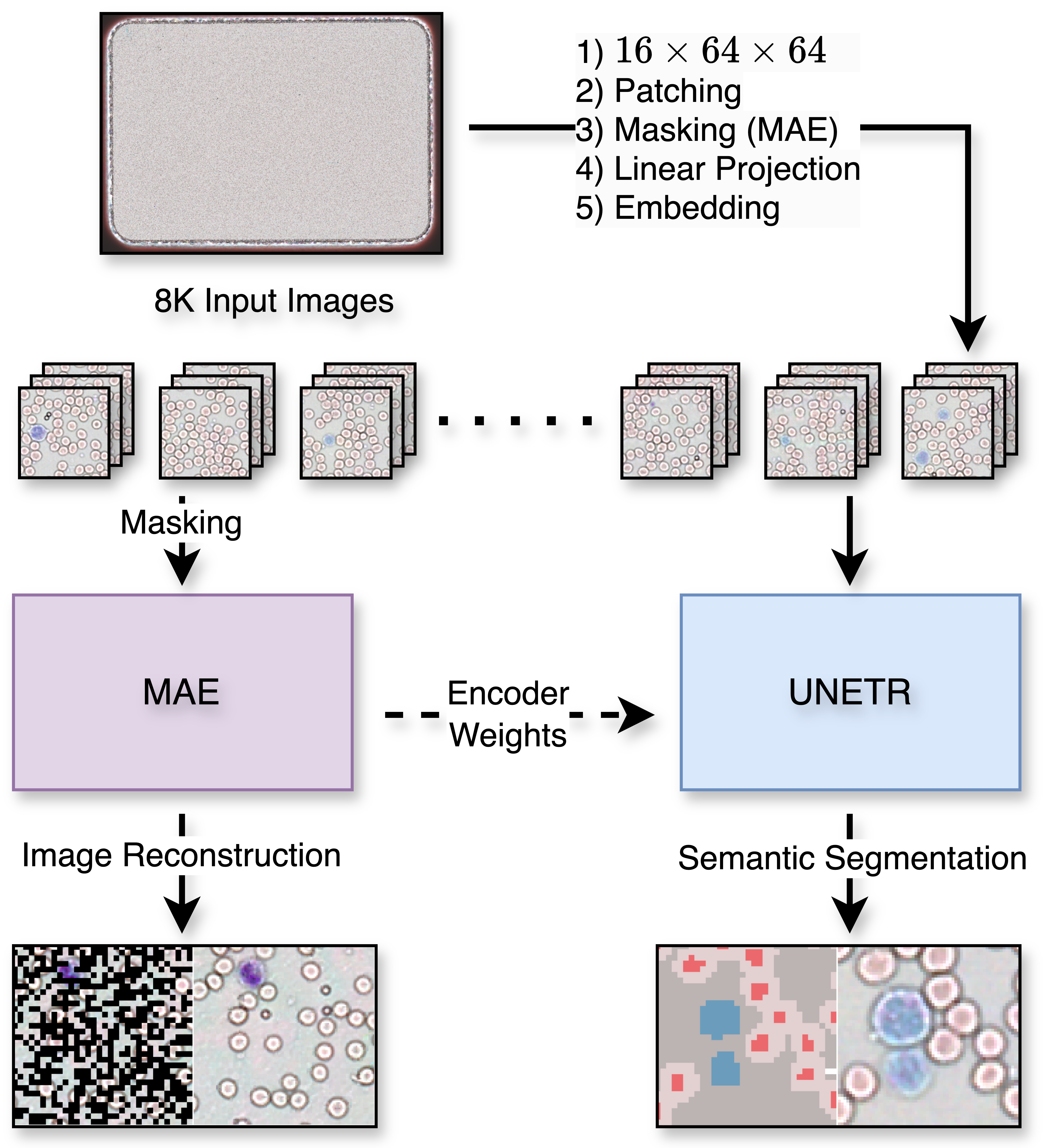}
    \vspace{-0.2cm}
    \caption{Overview of semantic segmentation framework. Input image was divided into smaller images of dimension $16 \times 64 \times 64$, after which patching and masking (MAE only) was applied. Subsequently, linear projection and embedding was performed, followed by training a MAE. ViT output encoder weights were then (optionally) used to trained a UNETR for pixel-wise classification to merge both local and global contexts.}
    \label{fig:overview}
\end{figure}

\FloatBarrier

Recently, several deep learning techniques for segmentation and classification of medical imagery have emerged~\cite{Saleem::2022a,Garcia-Lamont::2024a}.
For example, fully convolutional networks such as the U-Net architecture has been widely used for segmentation of medical images~\cite{Ronneberger::2015a,Livermore::2022a}.
The network consists of an encoder that extracts representations by sequential downsampling to a bottleneck feature representation.
A decoder is then used to reconstruct representations from the bottleneck structure by upsampling using transposed convolutions.
An important component of the U-Net architecture is the use of skip connections that merges each corresponding downsampled and upsampled feature representations.
This provides a mechanism to maintain spatial context at higher resolutions through the network.
The self-attention mechanism from the transformer architecture (\textit{i.e.}, multi-head self-attention) is another common approach that has been integrated as the key building block for numerous applications~\cite{Saleem::2022a,Manzari::2023a}.
In the medical imagery domain, the vision transformer (ViT) has attracted a lot of attention for classification, detection, and segmentation of human anatomy for diseases.
In part, this is due to the ability of the ViT architecture to capture long-range correlations to define a global context~\cite{Madan::2024a,Shamshad::2023b}.
The U-Net Transformer (UNETR) is an emerging technique~\cite{Hatamizadeh::2021a}, where the aforementioned approaches are combined that enables a network to capture and maintain local spatial contexts using convolutions, combined with long-range dependencies from a transformer at different resolutions
In particular, the approach uses a ViT encoder (\textit{e.g.}, optionally pre-trained) with skip connections with encoder outputs and/or intermediate outputs with the decoder of the U-Net architecture.

In this study, we used the Prospector point-of-care device from Alentic Microscience Inc.~\cite{Alentic.::2025a} to obtain high-resolution images of blood samples.
This device uses lensless on-chip near-field microscopy to provide 16-channel high-resolution images in less 10 minutes of acquisition time, containing red blood cells, white blood cells, and platelets that can be used for segmentation and classification.
In particular, we investigated a semantic segmentation framework (Fig.~\ref{fig:overview}) using a pre-trained ViT encoder from masked autoencoders (MAE) or randomly initialized UNETR networks.
We divided the input image into multiple images of dimension $16 \times 64 \times 64$ that were provided to a MAE for pre-training or directly to the UNETR network.
This divide-and-conquer approach drastically reduced the compute requirements for MAE networks and is particularly well-suited for small objects.
In addition, we reduced the embedding dimension, the number of self-attention heads, and the number of layers for the encoder.
Ideally, minimizing network size to reduce the on-device memory requirements. 
We investigated different patch sizes and for the MAE networks, we applied different mask ratios.

The main contributions of our work can be summarized as the following:

\begin{itemize}
    \item We propose an efficient semantic segmentation framework for small blood components using a customized UNETR network with or without a pre-trained ViT encoder.
    \item We show segmentation masks that demonstrate pixel-sized classification of platelets and aggregates.  
    \item We validated the effectiveness of our proposed methods by comparing to randomly initialized UNETR networks.
\end{itemize}

% RELATED
\section{Related Work}
\label{sect:related}
The semantic segmentation of small blood components, such as platelets is difficult because of:
1) small sizes; 
2) tendency to form aggregates; and
3) often indistinguishable features from background and surrounding objects.
As a result, a neural network must learn to extract representations for single entity objects that can be as small as a single pixel and/or exist as complex structures with or without other blood components.
There are limited examples of deep learning strategies for high-precision detection and classification of platelets.

The traditional approach for the preparation and analyses of blood samples involves a pathologist or qualified technician to generate blood smear samples, followed by image collection and processing, segmentation, classification, and manual counting.
While much of this process has now been automated to obtain patient data analytics with the use of flow cytometry and similar technologies, there continues a need for blood smear preparations to obtain detailed cellular-level information.
In particular, much attention has focused on the sub-classification of leukocytes.

Chowdhury \textit{et al.}~\cite{Chowdhury::2020a} reported the automated complete blood count using convolution neural networks (CNN) by creating a dataset for different blood smear data sources.
The authors assembled dataset consists of healthy and malaria infected red blood cells, platelets, platelet aggregates, and white blood cells and their subtypes: eosinophils, basophils, neutrophils, lymphocytes, and monocytes.
After data processing to obtain single-cell images along with their corresponding labels, the authors subsequently merged the single-cell images on $1000 \times 1000$ dimension background images.
The resulting images were then used to generate datasets containing $224 \times 224$ dimension images and corresponding labels.
The authors then applied scaled-down networks derived from the You Only Look Once (YOLO)~\cite{Redmon::2015a} and ResNet-18~\cite{He::2015a} neural networks.
Classification accuracy of 97.2\% was reported for the YOLO network, whereas for the ResNet-18 network, 94.7\% accuracy.
As compared to our study, we acquired 8K resolution images directly using the Prospector device, without the need of extensive manual data processing.
In addition, since we opted to use pixel-level classification (\textit{i.e.}, semantic segmentation), this allowed a flexible choice of input sizes for networks without resizing using interpolation techniques.

In 2020, Li \textit{et al.}~\cite{Li::2020a} investigated the semantic segmentation of white blood cells using a U-Net architecture.
In this case, input images were converted from the Red-Green-Blue (RGB) color space to the Hue-Saturation-Value (HSV) color space.
This was motivated from the observation that red blood cells and platelets can be differentiated easier in the HSV color space.
Following the color space conversion, the authors applied several preprocessing techniques (\textit{e.g.}, thresholding) to create datasets for training and evaluation that contained only white blood cells.
An additional class was added for the interface of more than one white blood cell; in addition, class weighting for background and for learning small separation boundaries of adjoining cells.
Using their U-Net model with data augmentation, the authors reported an accuracy of 97.9\% and an F1-score of 88.4\%. 

While these examples focused mostly on white blood cell and reported high performance scores, when evaluating platelets and/or aggregates, these scores (if reported) are reduced.
Platelets are difficult to classify because of their size and high variability (\textit{vide supra}).
Our main focus in this paper is the classification of platelets.
Accurate detection and quantification of these blood components would support healthcare professional to evaluate platelet related health risks.

% METHODS
\section{Methods}
\label{sect:methods}
\subsection{Instrumentation and Data Acquisition}

The device used to create the datasets for this study was the Prospector, a lensless near-field microscope developed by Alentic Microscience, Inc.~\cite{Alentic.::2025a}.
It contains a monochromatic sensor and an array of RGB and ultraviolet light-emitting diodes (LEDs).
These diodes are juxtaposed to illuminate a blood sample from multiple angles, to provide super-resolved images which are then converted to a 16-channel input used for this study.
Blood samples are prepared by applying a single drop of blood, mixed with reagents on the device sensor.
A consumable chambertop is subsequently affixed to generate a monolayer of blood components before capturing images.
High-resolution images are obtained using LEDs of the sensor from different angles to produce $12 \times$ RGB images of dimensions $3 \times 3K \times 4K$.
These images were then super-resolved ($1 \times$ RGB, super-resolved) resulting with a dimension of $3 \times 6K \times 8K$.
Finally, images were re-arranged to 12-channels and subsequently added orthogonal inputs from RGB and ultraviolet frames, providing input data with dimensions $16 \times 3K \times 4K$.
These were then processed and used for the creation of datasets describe in the next section.

\subsection{Datasets and Network Implementations}

\begin{figure}[!htb]
    \centering
    \subfigure
    {
        \includegraphics[width=0.23\textwidth]{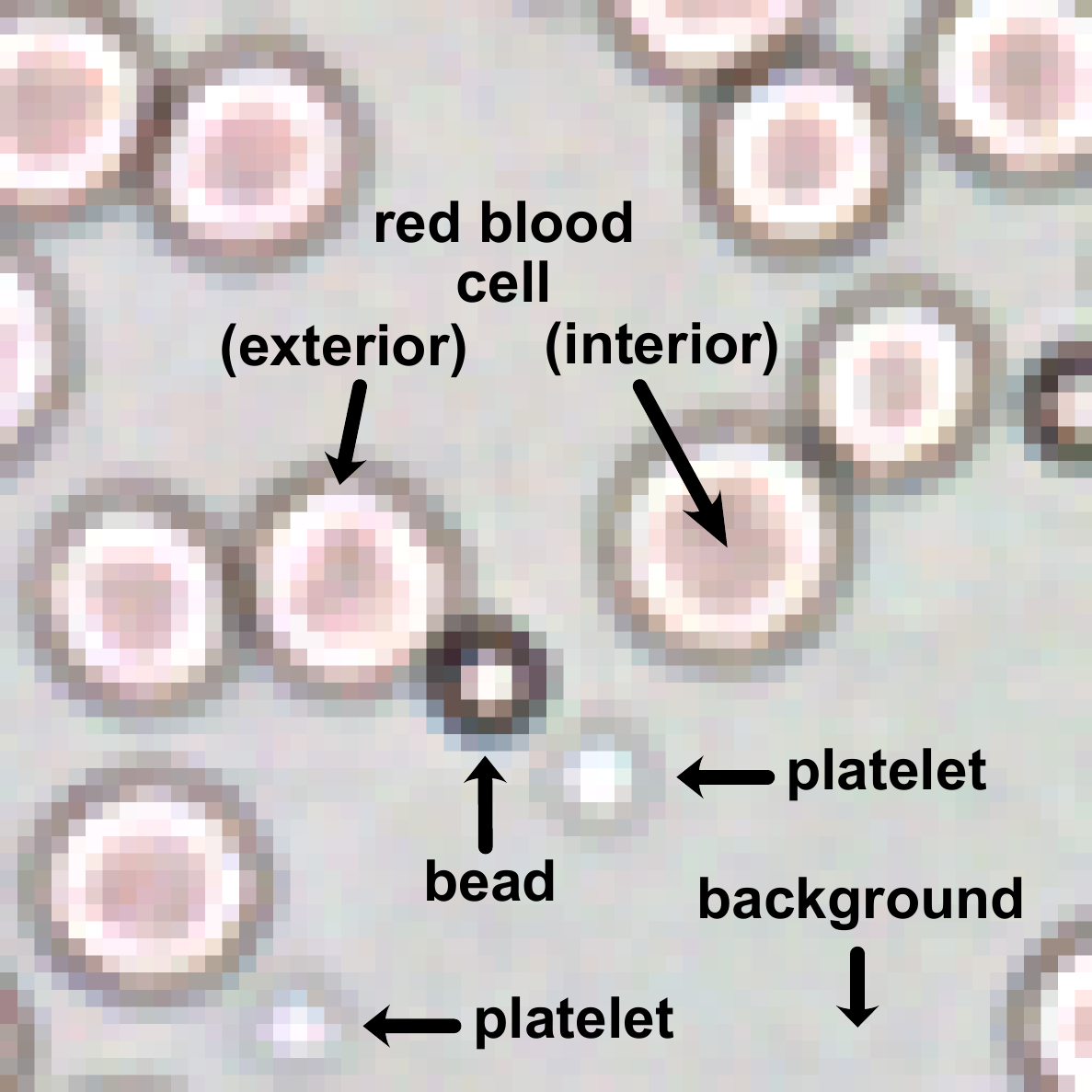}
        \label{fig:rbc}
    }
    \vspace{-0.2cm}
    \hspace{-0.4cm}
    \subfigure
    {
        \includegraphics[width=0.23\textwidth]{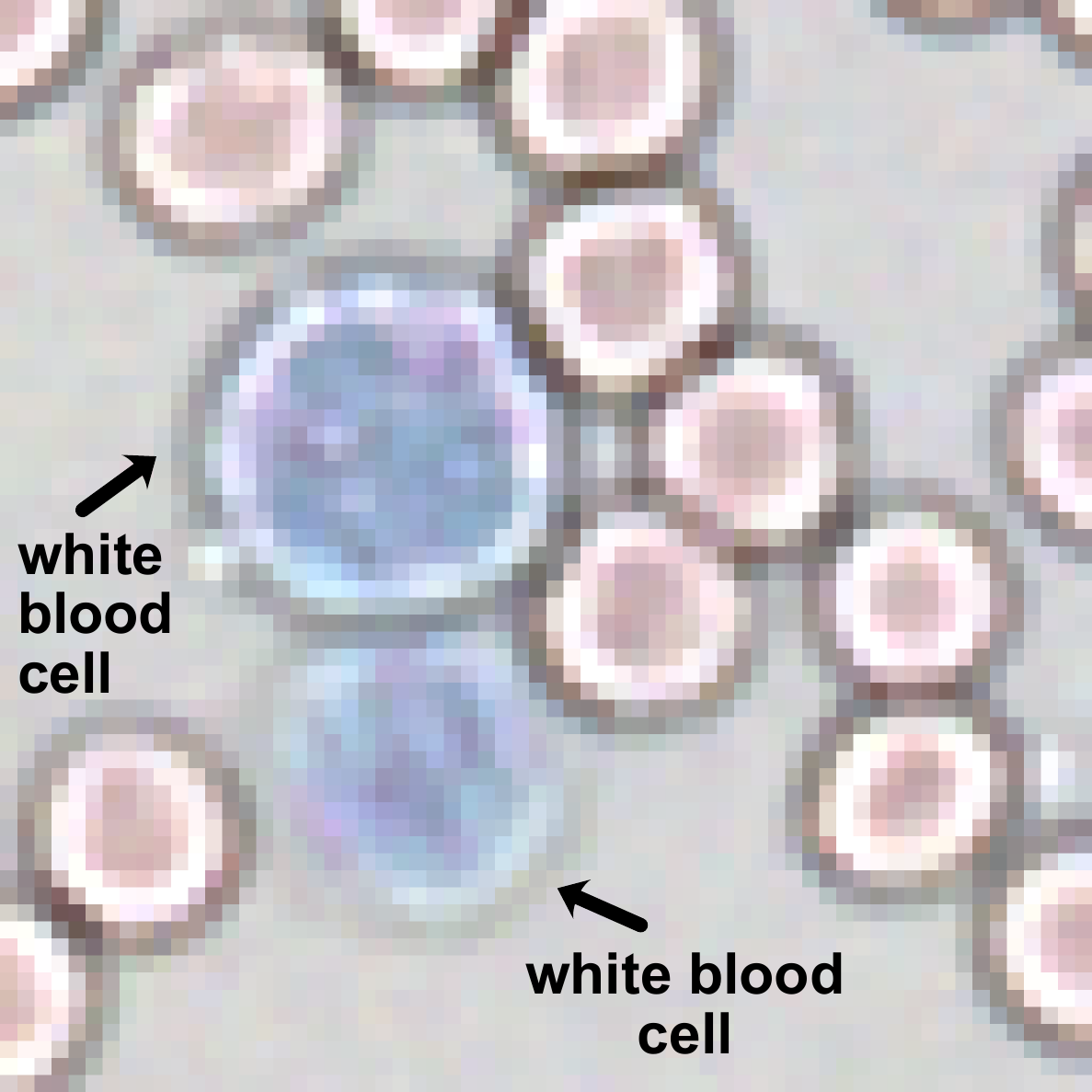}
        \label{fig:wbc}
    }
    \caption{Sample images with labeled classes. Artifacts, debris, and bubbles omitted for clarity.}
    \label{fig:class_samples}
\end{figure}

Images used consisted of multiple instances of one or more of the following classes:
1) background; 2) white blood cell (WBC); 3) platelet; 4) red blood cell (RBC, exterior); 5) red blood cell (RBC, interior); 6) bead; 7) artifacts; 8) debris; and 9) bubble (Fig.~\ref{fig:class_samples}).

\FloatBarrier

\noindent
In the following sections, descriptions for the preparation of datasets, network configurations, and hyperparameters are provided for self-supervised ViT-MAEs networks and semantic segmentation using UNETR networks. 

\subsubsection{Self-supervised masked autoencoders (ViT-MAE)}
The dataset used for mask autoencoders consists of 23,040 images of dimension $16 \times 256 \times 256$, derived from input data described above.
These were then partitioned into training (80\%), validation (10\%), and testing (10\%) datasets.
During training and inference, a collate function was used to further reduced the size of input to $16 \times 64 \times 64$.
Random horizontal and vertical flip data augmentations were applied during training.
This resulted with a training dataset with 294,912 images and 36,864 each for the validation and testing datasets.

A ViT-MAE for pre-training network was used from the Hugging Face transformers library~\cite{Hugging-Face::2025a}.
A symmetric encoder-decoder network was configured using embedding and multi-layer perceptron dimensions of 192, and 768, respectively.
In both cases, 6 encoder layers with 6 multi-head attention mechanisms were configured.
A layer normalization of $1.0 \times 10^{-6}$ was applied during training.
Networks were configured using different patch sizes (2, 4, or 8) and mask ratios (0.5, 0.75, or 0.9).
All other values were used with there default settings.
Training was done over 100 epochs (batch size 16), using the Adam optimizer ($\beta_1 = 0.9, \beta_2 = 0.95$) with a learning rate of $1.0 \times 10^{-3}$.
The OneCycleLR~\cite{Smith::2017a,PyTorch::2025a} scheduler was applied during training with an initial and final learning rate factor of 0.01, where the rate was increased over the first 10\% of the total number of epochs. 
Mean absolute (reconstruction) error was computed using the available function from the transformers library with normalization of pixel values.

\subsubsection{Semantic segmentation (UNETR)}
The image dataset used for semantic segmentation consists of 2382 images with input dimensions $16 \times 64 \times 64$ and corresponding sparse pixel-wise labels with dimensions $64 \times 64$.
The dataset contained the following respective class pixel counts:
background (80,166), 
white blood cells (7442), 
platelets (1472), 
red blood cells (exterior, 14,506), 
red blood cells (interior, 2422), 
beads (749), 
artifacts (5,827), 
debris (62,161), and 
bubbles (49,966).
This represents 2.3\% of the 2383 $\times$ 64 $\times$ 64 (9,760,768) total pixels for labeled data.
The dataset was then used for 5-fold cross-validations using random horizontal and vertical flip data augmentations during training with a unique seed to ensure the same transformation to both the image and label are applied.
Three variants of the UNTER network~\cite{Hatamizadeh::2021a} were configured using classes from the UNETR  Block available from the MONAI framework~\cite{MONAI-Consortium::2020a} with a spatial dimension of 2 and 16 input channels.
For semantic segmentation, different feature sizes (16, 32, 64, 128, 256, or 512) and patch sizes (2, 4, or 8) (using randomly initialized or with pre-trained encoders from ViT-MAEs) were investigated.
For patch size 2, hidden features from layer 6 (encoder output) was used for decoder upsampling and skip connections; patch size 4, hidden features from layer 6 and hidden features from layer 3; and for patch size 4, hidden features from layer 6 and hidden features from layers 2 and 4.
Training was done using 5-fold cross-validation over 200 epochs (batch size 64), and the Adam optimizer with a learning rate of $1.0 \times 10^{-3}$.
The OneCycleLR scheduler was applied during training with an initial and final learning rate factor of 0.01, where the rate was increased over the first 10 epochs and decayed to 100 epochs, where the learning rate was then maintained at $1.0 \times 10^{-5}$.
In the case of pre-trained encoders, encoder weights were fixed for the first 100 epochs.

%%%%%%%%%%%%%%%%%%%%%%%%%%%%%%%%%%%%
\begin{table*}[!htb]
\begin{adjustbox}{width=0.98\textwidth}
    \centering
    \begin{threeparttable}
        \caption{Downstream application with patch sizes 2, 4, and 8 for semantic segmentation using randomly initialized UNETR networks.\tnote{1,2}}
        \begin{tabular}{ l l l l l l l l l }
            \toprule
            % headings
            \textbf{Patch Size}
            & \textbf{Feature Size}
            & \textbf{Accuracy}
            & \textbf{F1-score}
            & \multicolumn{5}{c}{\textbf{F1-score}} \\
            \cmidrule(lr){5-9}
            & 
            & 
            & 
            & \textbf{Platelet} 
            & \textbf{WBC} 
            & \textbf{RBC (interior)} 
            & \textbf{RBC (exterior)} 
            & \textbf{Bead}
            \\
            \midrule
            % entry: logs/crossunetr/2_512
            2
            & 512
            & $0.980 \pm 0.005$
            & $0.947 \pm 0.014$
            & $0.924 \pm 0.008$
            & $0.959 \pm 0.017$
            & $0.943 \pm 0.010$
            & $0.978 \pm 0.005$
            & $0.974 \pm 0.014$
            \\
            % entry: logs/crossunetr/2_256
            2
            & 256
            & $0.981 \pm 0.004$
            & $0.945 \pm 0.014$
            & $0.921 \pm 0.010$
            & $0.960 \pm 0.015$
            & $0.941 \pm 0.012$
            & $0.978 \pm 0.004$
            & $0.971 \pm 0.010$
            \\
            % entry: logs/crossunetr/2_128
            2
            & 128
            & $0.982 \pm 0.007$
            & $0.942 \pm 0.021$
            & $0.907 \pm 0.012$
            & $0.959 \pm 0.017$
            & $0.937 \pm 0.007$
            & $0.979 \pm 0.004$
            & $0.972 \pm 0.015$
            \\
            % entry: logs/crossunetr/2_64
            2
            & 64
            & $0.980 \pm 0.007$
            & $0.936 \pm 0.017$
            & $0.898 \pm 0.008$ 
            & $0.954 \pm 0.019$
            & $0.928 \pm 0.012$
            & $0.973 \pm 0.005$
            & $0.954 \pm 0.017$
            \\
            % entry: logs/crossunetr/2_32
            2
            & 32
            & $0.975 \pm 0.007$
            & $0.913 \pm 0.019$
            & $0.856 \pm 0.021$
            & $0.945 \pm 0.013$
            & $0.914 \pm 0.011$
            & $0.966 \pm 0.005$
            & $0.929 \pm 0.021$
            \\
            % entry: logs/crossunetr/2_16
            2
            & 16
            & $0.954 \pm 0.015$
            & $0.819 \pm 0.019$
            & $0.736 \pm 0.024$
            & $0.908 \pm 0.023$
            & $0.850 \pm 0.028$
            & $0.940 \pm 0.009$
            & $0.762 \pm 0.053$
            \\
            \midrule
            % entry: logs/crossunetr/4_512
            4
            & 512
            & $0.982 \pm 0.005$
            & $0.948 \pm 0.014$
            & $0.921 \pm 0.008$
            & $0.958 \pm 0.016$
            & $0.943 \pm 0.011$
            & $0.979 \pm 0.004$
            & $0.974 \pm 0.009$
            \\
            % entry: logs/crossunetr/4_256
            4
            & 256
            & $0.982 \pm 0.004$
            & $0.946 \pm 0.013$
            & $0.913 \pm 0.010$
            & $0.961 \pm 0.015$
            & $0.941 \pm 0.009$
            & $0.973 \pm 0.017$
            & $0.972 \pm 0.006$
            \\
            % entry: logs/crossunetr/4_128
            4
            & 128
            & $0.981 \pm 0.007$
            & $0.942 \pm 0.019$
            & $0.909 \pm 0.011$
            & $0.957 \pm 0.014$
            & $0.937 \pm 0.012$
            & $0.973 \pm 0.014$
            & $0.963 \pm 0.021$
            \\
            % entry: logs/crossunetr/4_64
            4
            & 64
            & $0.979 \pm 0.006$
            & $0.931 \pm 0.016$
            & $0.890 \pm 0.010$
            & $0.951 \pm 0.021$
            & $0.927 \pm 0.012$
            & $0.972 \pm 0.005$
            & $0.957 \pm 0.011$
            \\
            % entry: logs/crossunetr/4_32
            4
            & 32
            & $0.975 \pm 0.006$
            & $0.911 \pm 0.010$
            & $0.853 \pm 0.017$
            & $0.943 \pm 0.015$
            & $0.912 \pm 0.019$
            & $0.967 \pm 0.003$
            & $0.919 \pm 0.022$
            \\
            % entry: logs/crossunetr/4_16
            4
            & 16
            & $0.961 \pm 0.007$
            & $0.823 \pm 0.026$
            & $0.714 \pm 0.067$
            & $0.924 \pm 0.021$
            & $0.849 \pm 0.023$
            & $0.945 \pm 0.011$
            & $0.826 \pm 0.059$
            \\
            \midrule
            % entry: logs/crossunetr/8_512
            8
            & 512
            & $0.982 \pm 0.005$
            & $0.946 \pm 0.013$
            & $0.918 \pm 0.005$
            & $0.959 \pm 0.016$
            & $0.940 \pm 0.009$
            & $0.979 \pm 0.004$
            & $0.971 \pm 0.015$
            \\
            % entry: logs/crossunetr/8_256
            8
            & 256
            & $0.980 \pm 0.006$
            & $0.943 \pm 0.015$
            & $0.918 \pm 0.008$
            & $0.960 \pm 0.016$
            & $0.940 \pm 0.011$
            & $0.978 \pm 0.004$
            & $0.962 \pm 0.012$
            \\
            % entry: logs/crossunetr/8_128
            8
            & 128
            & $0.981 \pm 0.006$
            & $0.942 \pm 0.018$
            & $0.916 \pm 0.012$
            & $0.961 \pm 0.012$
            & $0.941 \pm 0.013$
            & $0.977 \pm 0.004$
            & $0.963 \pm 0.016$
            \\
            % entry: logs/crossunetr/8_64
            8
            & 64
            & $0.989 \pm 0.006$
            & $0.930 \pm 0.020$
            & $0.894 \pm 0.015$
            & $0.958 \pm 0.011$
            & $0.931 \pm 0.016$
            & $0.972 \pm 0.005$
            & $0.947 \pm 0.017$
            \\
            % entry: logs/crossunetr/8_32
            8
            & 32
            & $0.974 \pm 0.010$
            & $0.909 \pm 0.029$
            & $0.860 \pm 0.015$
            & $0.946 \pm 0.017$
            & $0.909 \pm 0.021$
            & $0.965 \pm 0.004$
            & $0.934 \pm 0.021$
            \\
            % entry: logs/crossunetr/8_16
            8
            & 16
            & $0.960 \pm 0.012$
            & $0.823 \pm 0.021$
            & $0.721 \pm 0.026$
            & $0.921 \pm 0.012$
            & $0.852 \pm 0.024$
            & $0.940 \pm 0.015$
            & $0.838 \pm 0.018$
            \\
            \midrule
            \bottomrule
        \end{tabular}
        $^1$  Models used randomly initialized weights for both encoder and decoder.
        $^2$ Metrics represents 5-fold cross-validation providing mean values along with their corresponding standard deviations.
        Values for accuracy and F1-scores represents all 9 classes.
        Individual background, artifacts, debris, and bubbles omitted for clarity.
        \label{tab:unetr_random}
    \end{threeparttable}
\end{adjustbox}
\end{table*}

\FloatBarrier

% EXPERIMENTS
\section{Experiments}
\label{sect:experiments}
\subsection{Semantic Segmentation using a Randomly Initialized UNETR Network.}

To obtain baseline results, we initially trained randomly initialized UNETR networks with patch sizes of 2, 4, and 8.
In addition, we explored feature sizes from 6 to 512 for the embedding dimension of the input image used to merge with upsampled representations from the encoder.
These settings spanned a range of trainable network parameters from \textit{ca.} 3-200 million.
In Table~\ref{tab:unetr_random}, the 5-fold cross-validation results for patch sizes 2, 4, and 8 and feature sizes provided overall mean accuracy scores of 95.4--98.2\% and F1-scores of 82.3--94.8\%.
As we increased feature sizes, the results quickly plateaued at feature size 64 or 128.
Equally interesting was the minimal difference for the selection of different patch sizes.
Recall that for patch size 2, we only use the encoder output with the decoder, whereas for patch sizes 4 and 8 we additionally use hidden features from layer 3 or at layers 2 and 4, respectively.
Hence, using larger patch sizes result with larger networks and implies additional long-range contextual information from ViT encoders.
However, these results suggests that there is minimal benefit of increasing patch size for a given feature size.

\subsection{Pre-training an UNETR Encoder using ViT-MAEs}

In the previous section we presented the results from using randomly initialized networks; however, it is well-known that incremental performance can be realized using transfer learning by loading pre-trained weights.
For UNETR networks, we therefore seek to obtain pre-trained ViT encoders and in this domain, masked autoencoder have provided remarkable success for many applications~\cite{Huang::2023a}.
Table~\ref{tab:vitmae} shows the results

\begin{table}[!htb]
\begin{adjustbox}{width=0.48\textwidth} 
    \centering
    \begin{threeparttable}
        \caption{Network configurations, hyperparameters and loss metrics for ViT masked autoencoders.}
        \begin{tabular}{ l l l l l l }
            \toprule
            % headings
            \textbf{Model}\tnote{1}
            & \textbf{Mask Ratio}
            & \textbf{Patch Size}
            & \textbf{Epochs}
            & \textbf{Batch Size}
            & \textbf{Loss}
            \\
            \midrule
            % entry: logs/vitmae-random/02
            A1
            & 0.5
            & 2
            & 100
            & 16
            & 0.328
            \\
            % entry: logs/vitmae-random/04
            A2
            & 0.5
            & 4
            & 100
            & 16
            & 0.375
            \\
            % entry: logs/vitmae-random/06
            A3
            & 0.5
            & 8
            & 100
            & 16
            & 0.478
            \\
            % entry: logs/vitmae-random/08
            B1
            & 0.75
            & 2
            & 100
            & 16
            & 0.397
            \\
            % entry: logs/vitmae-random/10
            B2
            & 0.75
            & 4
            & 100
            & 16
            & 0.479
            \\
            % entry: logs/vitmae-random/12
            B3
            & 0.75
            & 8
            & 100
            & 16
            & 0.600
            \\
            % entry: logs/vitmae-random/14
            C1
            & 0.9
            & 2
            & 100
            & 16
            & 0.511
            \\
            % entry: logs/vitmae-random/16
            C2
            & 0.9
            & 4
            & 100
            & 16
            & 0.626
            \\
            % entry: logs/vitmae-random/18
            C3
            & 0.9
            & 8
            & 100
            & 16
            & 0.720
            \\
            \midrule
            \bottomrule
        \end{tabular}
        $^1$ Models are defined using encoders and decoders with [number of layers : number of heads : hidden size : intermediate size] as~$[6:6:192:768]$.
        Normalized pixels were used for loss calculations. 
        Unless otherwise specified, all other values were used as default settings provided from the Hugging Face transformers library for ViT-MAE configurations. 
        \label{tab:vitmae} 
    \end{threeparttable}
\end{adjustbox}
\end{table}

\begin{figure*}[!htb]
    \centering
    \subfigure[Mask ratio 0.5; patch size 2.]
    {
        \includegraphics[width=0.32\textwidth]{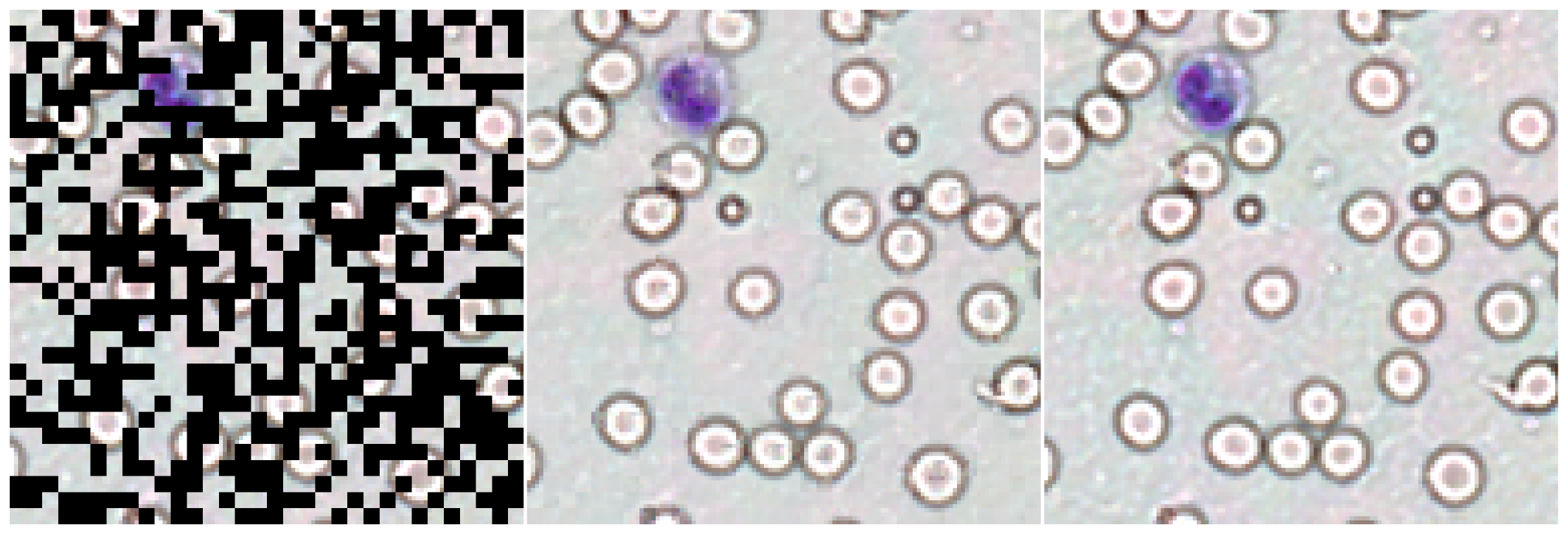}
        \label{fig:vitmae_random_02}
    }
    \vspace{-0.2cm}
    \hspace{-0.4cm}
    \subfigure[Mask ratio 0.5; patch size 4.]
    {
        \includegraphics[width=0.32\textwidth]{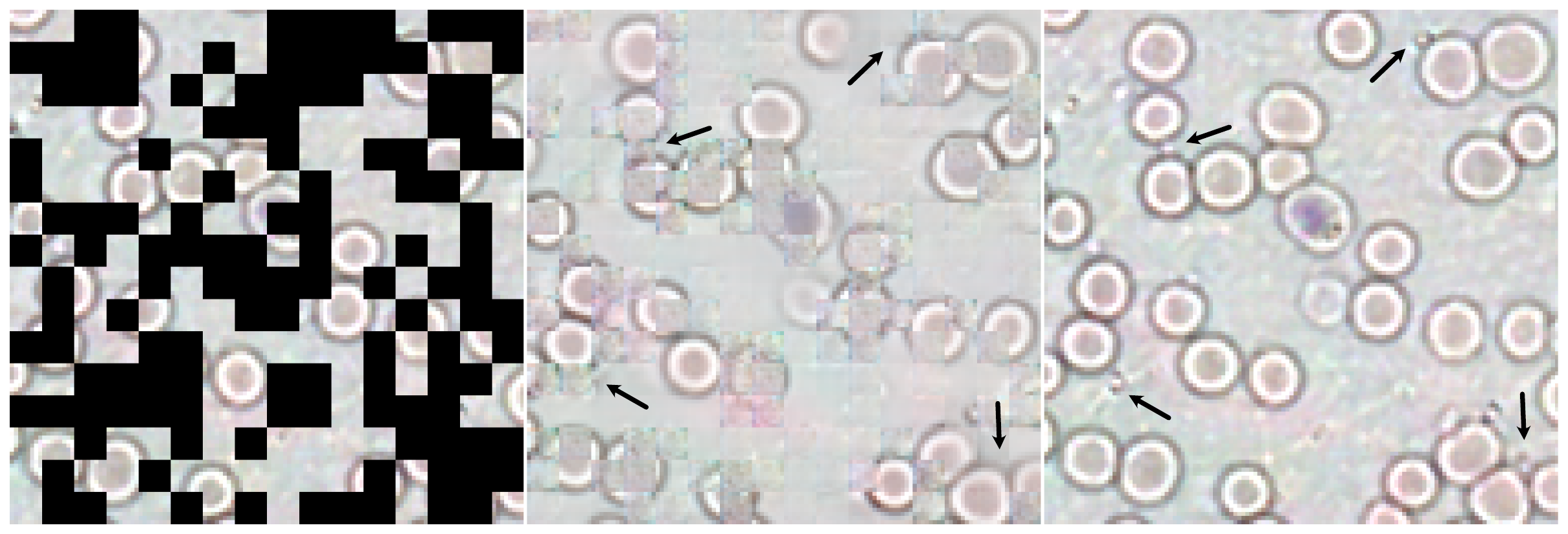}
        \label{fig:vitmae_random_04}
    }
    \hspace{-0.4cm}
    \subfigure[Mask ratio 0.5; patch size 8.]
    {
        \includegraphics[width=0.32\textwidth]{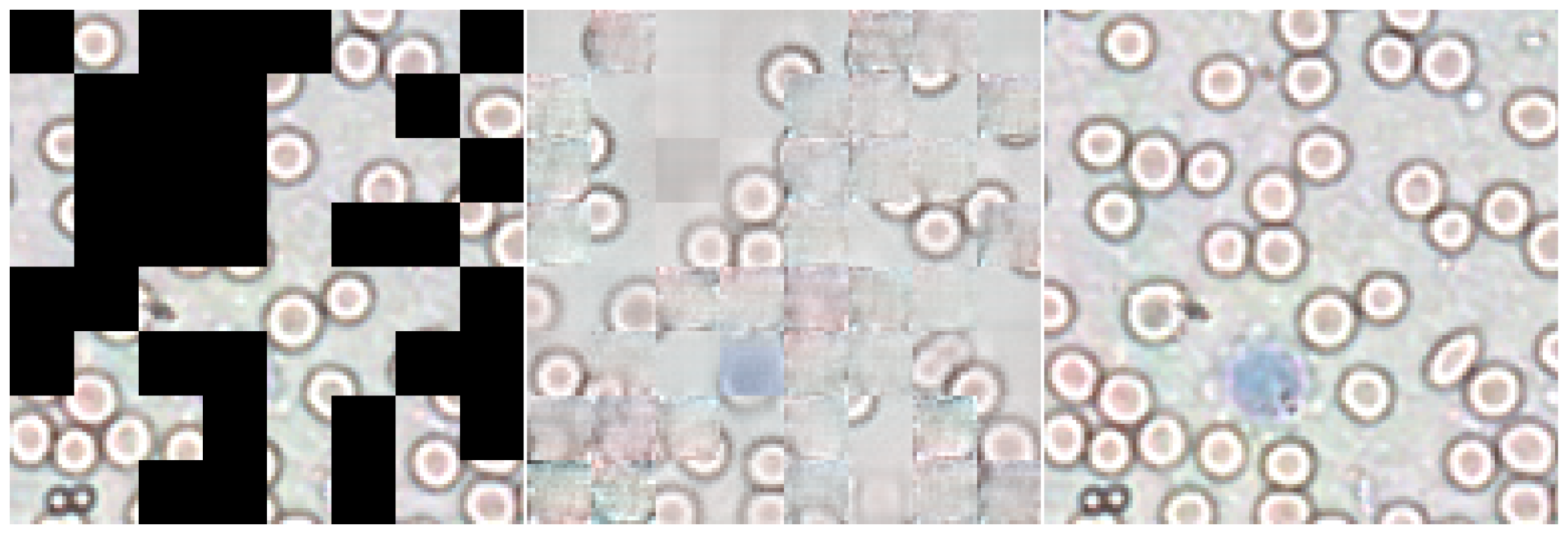}
        \label{fig:vitmae_random_06}
    }
    \subfigure[Mask ratio 0.75; patch size 2.]
    {
        \includegraphics[width=0.32\textwidth]{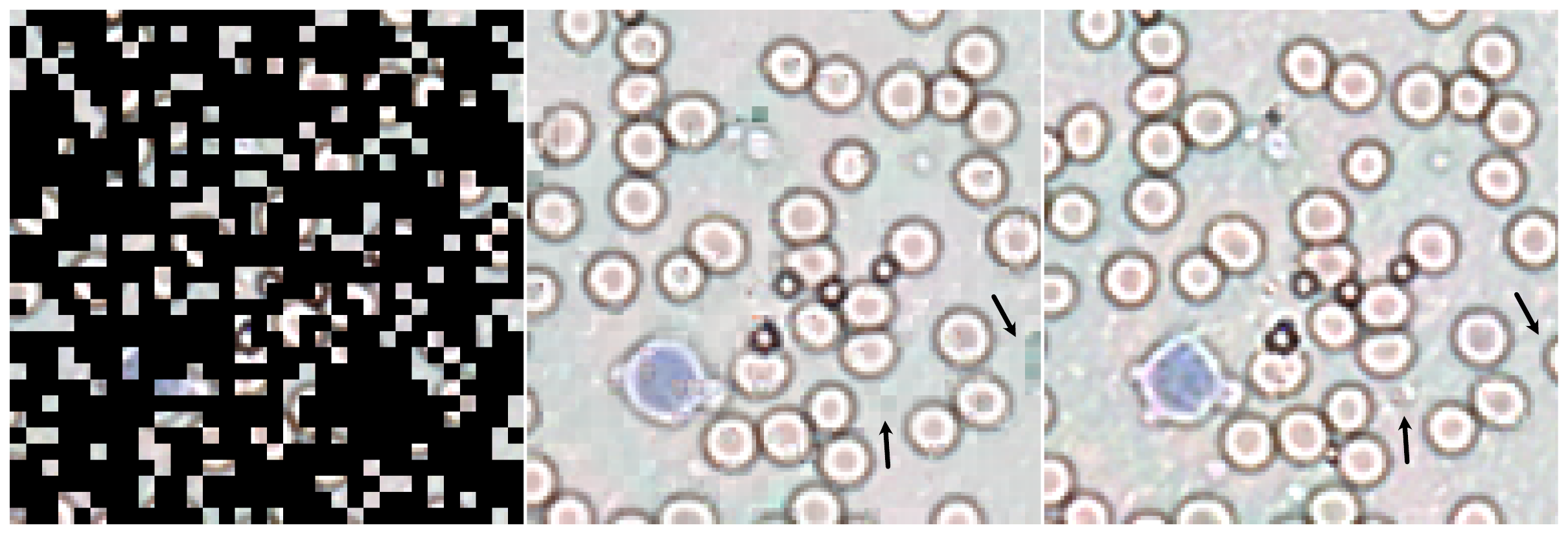}
        \label{fig:vitmae_random_08}
    }
    \hspace{-0.4cm}
    \vspace{-0.2cm}
    \subfigure[Mask ratio 0.75; patch size 4.]
    {
        \includegraphics[width=0.32\textwidth]{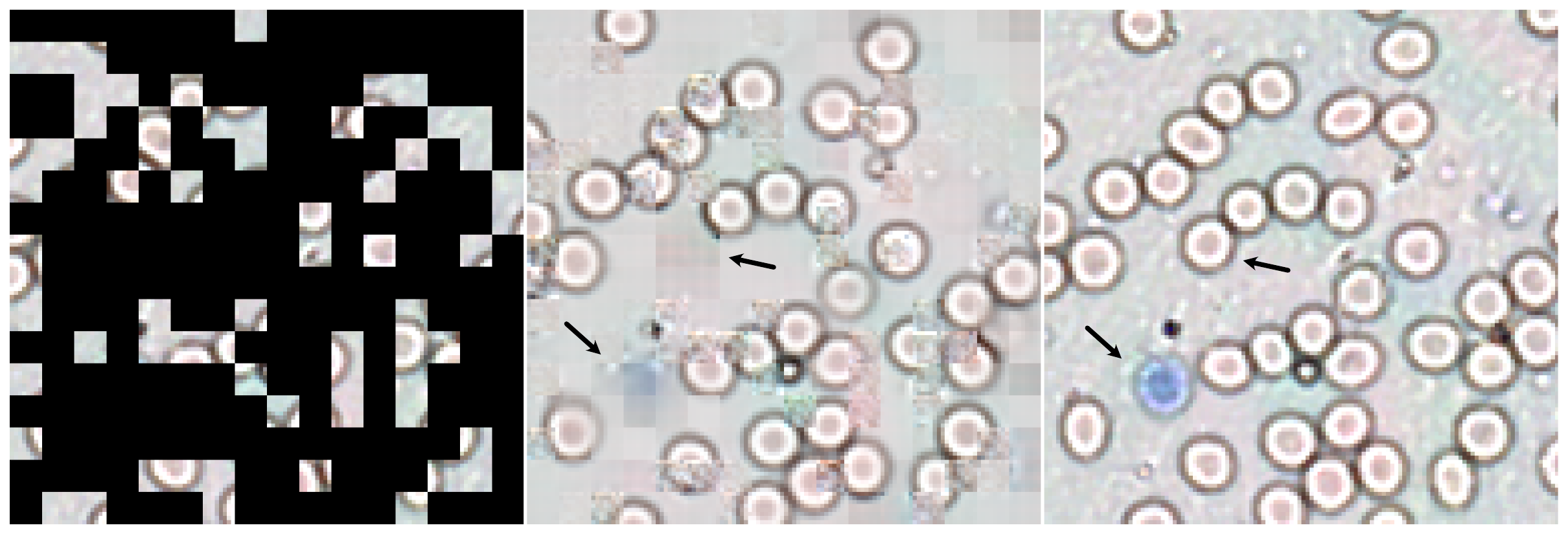}
        \label{fig:vitmae_random_10}
    }
    \hspace{-0.4cm}
    \subfigure[Mask ratio 0.75; patch size 8.]
    {
        \includegraphics[width=0.32\textwidth]{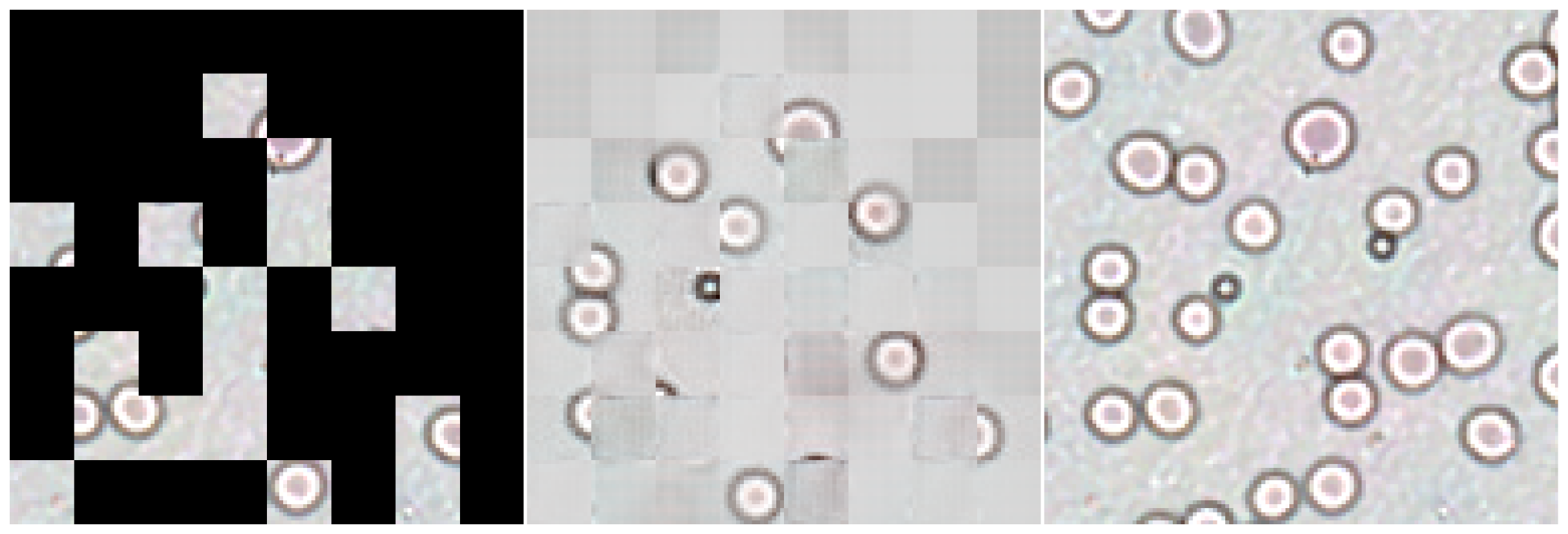}
        \label{fig:vitmae_random_12}
    }
    \subfigure[Mask ratio 0.9; patch size 2.]
    {
        \includegraphics[width=0.32\textwidth]{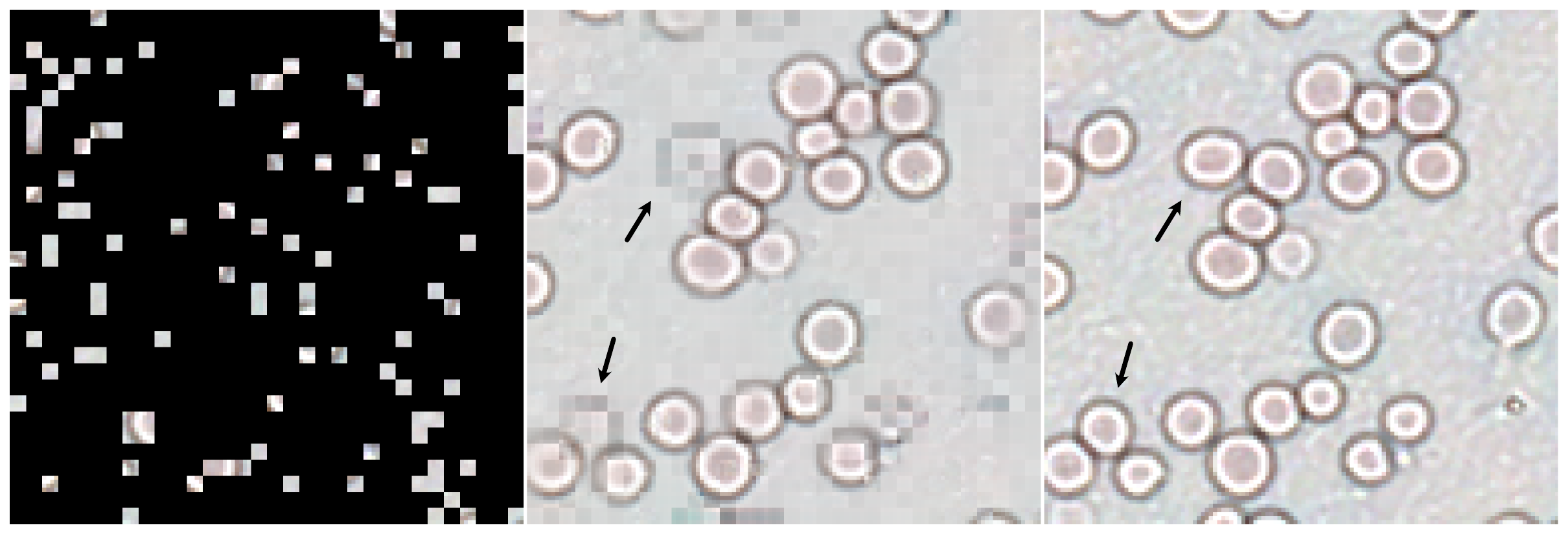}
        \label{fig:vitmae_random_14}
    }
    \hspace{-0.4cm}
    \subfigure[Mask ratio 0.9; patch size 4.]
    {
        \includegraphics[width=0.32\textwidth]{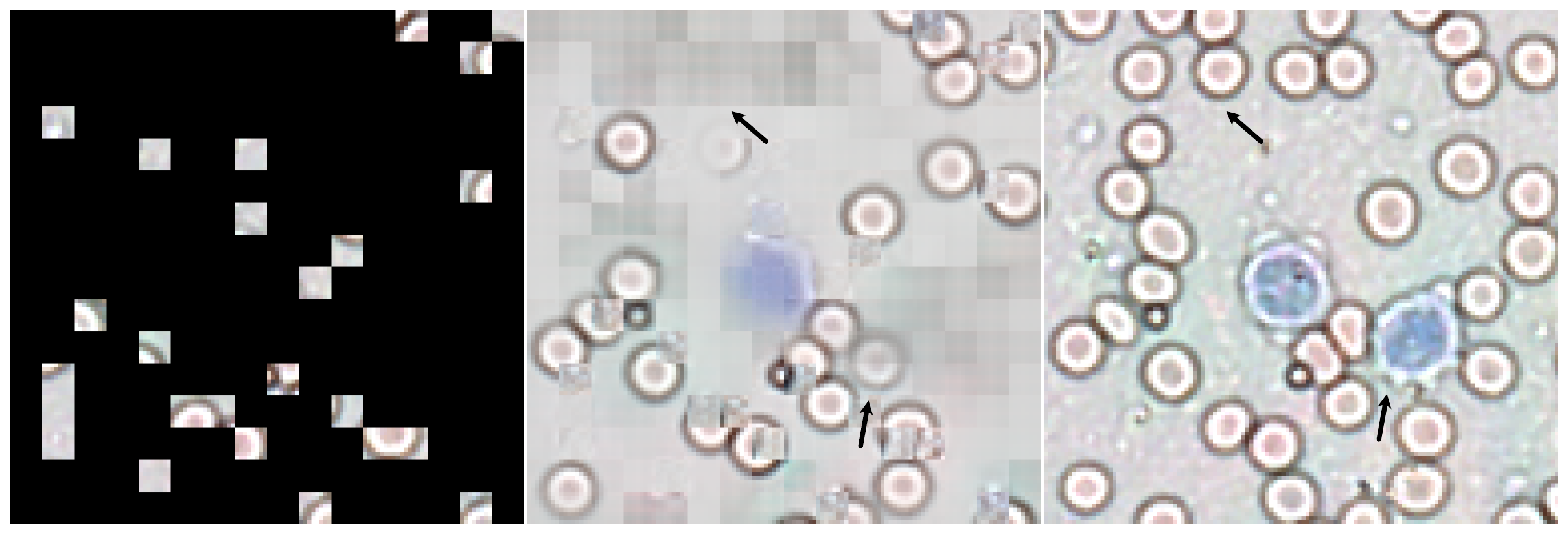}
        \label{fig:vitmae_random_16}
    }
    \hspace{-0.4cm}
    \subfigure[Mask ratio 0.9; patch size 8.]
    {
        \includegraphics[width=0.32\textwidth]{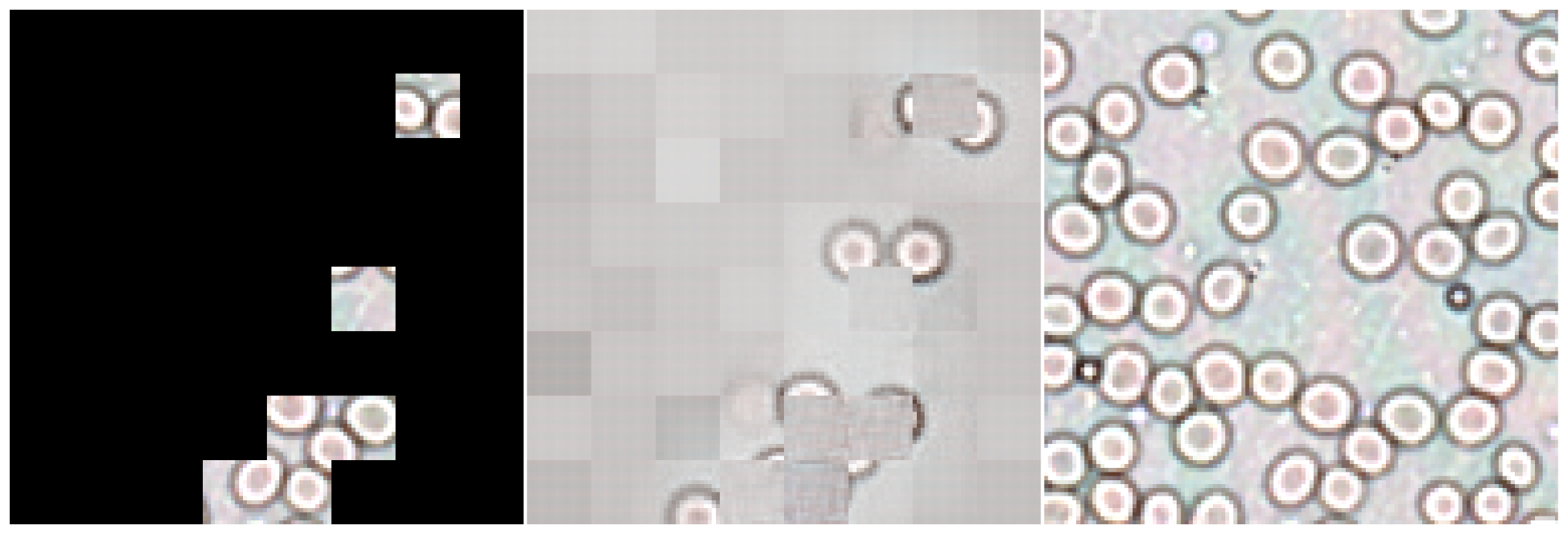}
        \label{fig:vitmae_random_18}
    }
    \vspace{-0.2cm}
    \caption{MAE reconstructions of super-resolved images. First row: Mask ratio 0.5; Second row: Mask ratio 0.75; and Third row: Mask ratio 0.9.
    Each row consists of plots for patch sizes 2, 4, and 8 (left-to-right), with masked images, reconstruction images, and original images (left-to-right). Black arrows compare objects in original image (right) to reconstructed image (center).
    }
    \label{fig:mae_results}
\end{figure*}

\FloatBarrier

\noindent
for training ViT-MAEs using different patch sizes and mask ratios.
We investigated patch sizes of 2, 4, and 8 to align with our previous randomly initialized UNETR networks.
In addition, we varied the mask ratios using values of 0.5, 0.75, and 0.9.
Increasing the mask ratios forces the networks to reconstruct more of the image space.
We provide the reconstructed images (Fig.~\ref{fig:mae_results}) here as indicator of performance; however, downstream applications of learned encoder weights would provide a quantitative evaluation.
Figure~\ref{fig:mae_results} shows the reconstruction images (center image of each panel) for the different pairings of patch size and mask ratio used for this study.
Based on these images, a mask ratio of 0.5 and patch size 2 (Fig.~\ref{fig:vitmae_random_02}) appeared to be optimal as a near perfect reconstruction was observed.

\FloatBarrier

\noindent
Using a mask ratio of 0.75 (Fig.~\ref{fig:vitmae_random_08}) resulted with similar outcomes but reconstruction discrepancies started to appear (noted by black arrows).
For all other combinations of mask ratios and patch sizes, additional discrepancies appeared such as missing cells or hallucinations (Fig.~\ref{fig:vitmae_random_16}, red blood cell \textit{vs.} white blood cell).
We therefore hypothesized that the ideal mask ratio and patch size would be 0.5 and 2, respectively.

\subsection{Semantic Segmentation using Pre-trained ViT Encoders and the UNETR Network.}

\begin{figure}[!htb]
    \centering
    \includegraphics[width=0.48\textwidth]{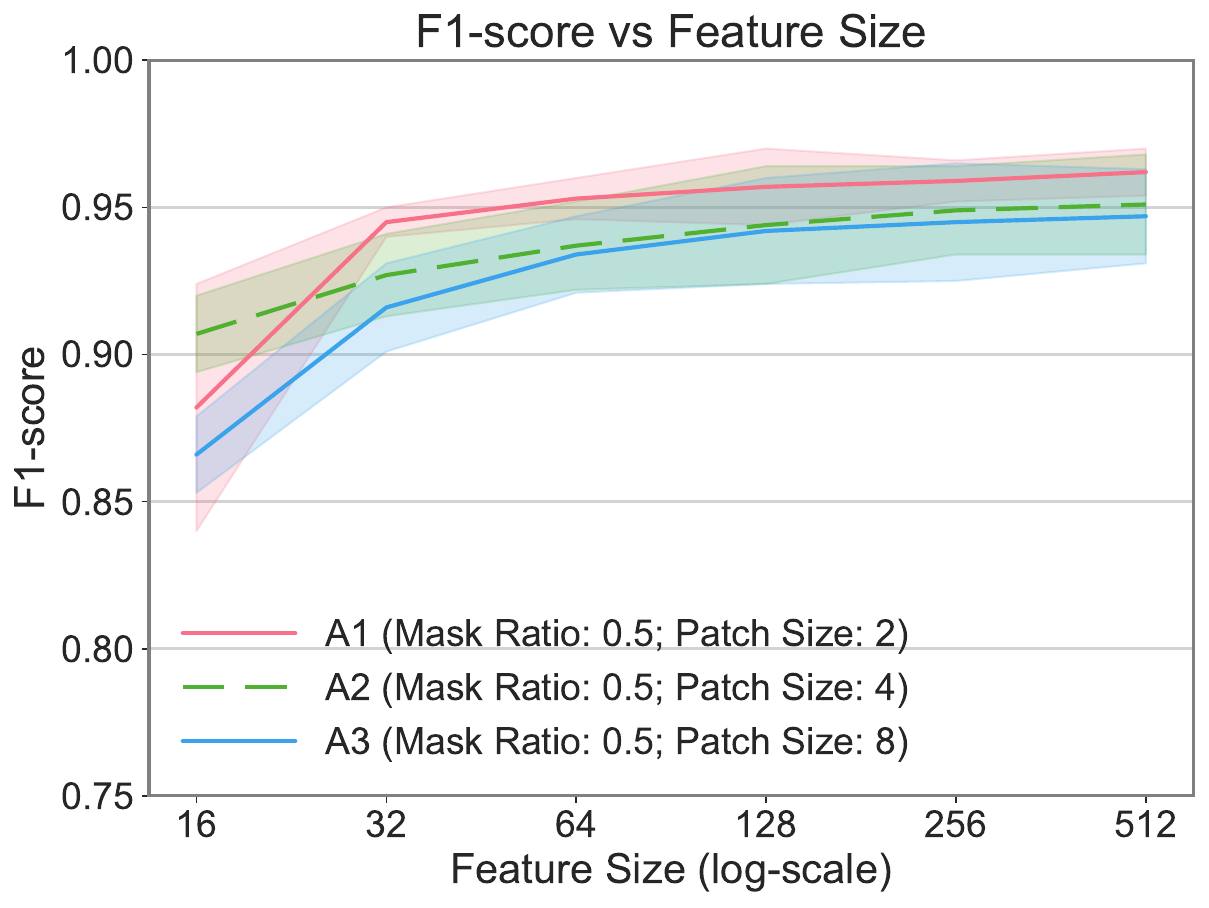}
    \vspace{-0.4cm}
    \caption{F1-scores for semantic segmentation with different feature sizes using pre-trained encoders A1--A3.
    Shaded areas represent respective standard deviations.}
    \label{fig:f1_05_2_4_8}
\end{figure}

To evaluate the performance of using randomly initialized UNETR network with those using pre-trained encoder, we investigated 9 possible configurations of mask ratios and patch sizes.
In particular, we applied 5-fold cross-validations to address the stochastic nature of training neural networks.
The results are displayed in Table~\ref{tab:unetr_combined}.
Recall that mask ratios are only relevant for the pre-training of the ViT encoders; therefore, for comparisons we can evaluate all mask ratios for a given patch size to the corresponding randomly initialized networks in Table~\ref{tab:unetr_random}.
While accuracy scores are provided, we discuss F1-scores given the unbalanced nature of the datasets.
For example, when using patch size 2 and random networks, mean F1-scores of 81.9--94.7\% were obtained for different feature sizes.
When using pre-trained encoders, 84.2--96.2\% were obtained for mean F1-scores.
This suggests minimal benefits of using pre-trained encoders; however, when considering their corresponding standards deviations, many of these are not statistically significant.
For example, Fig.~\ref{fig:f1_05_2_4_8} shows the results for the semantic segmentation using pre-trained encoders derived from A1, A2, and A3 with a mask ratio of 0.5 and patch sizes 2, 4, and 8 were used.
In this case, there is no statistical difference for the outcome of varying feature sizes with respect to F1-scores.

\FloatBarrier

\noindent
Although, when using patch size 2 and feature sizes 32 or 64, we observed slightly better performance for pre-trained encoder A1 as compared to pre-trained encoder A3 using patch size 8.

\begin{figure*}[!htb]
    \centering
    \subfigure[]
    {
        \includegraphics[width=0.32\textwidth]{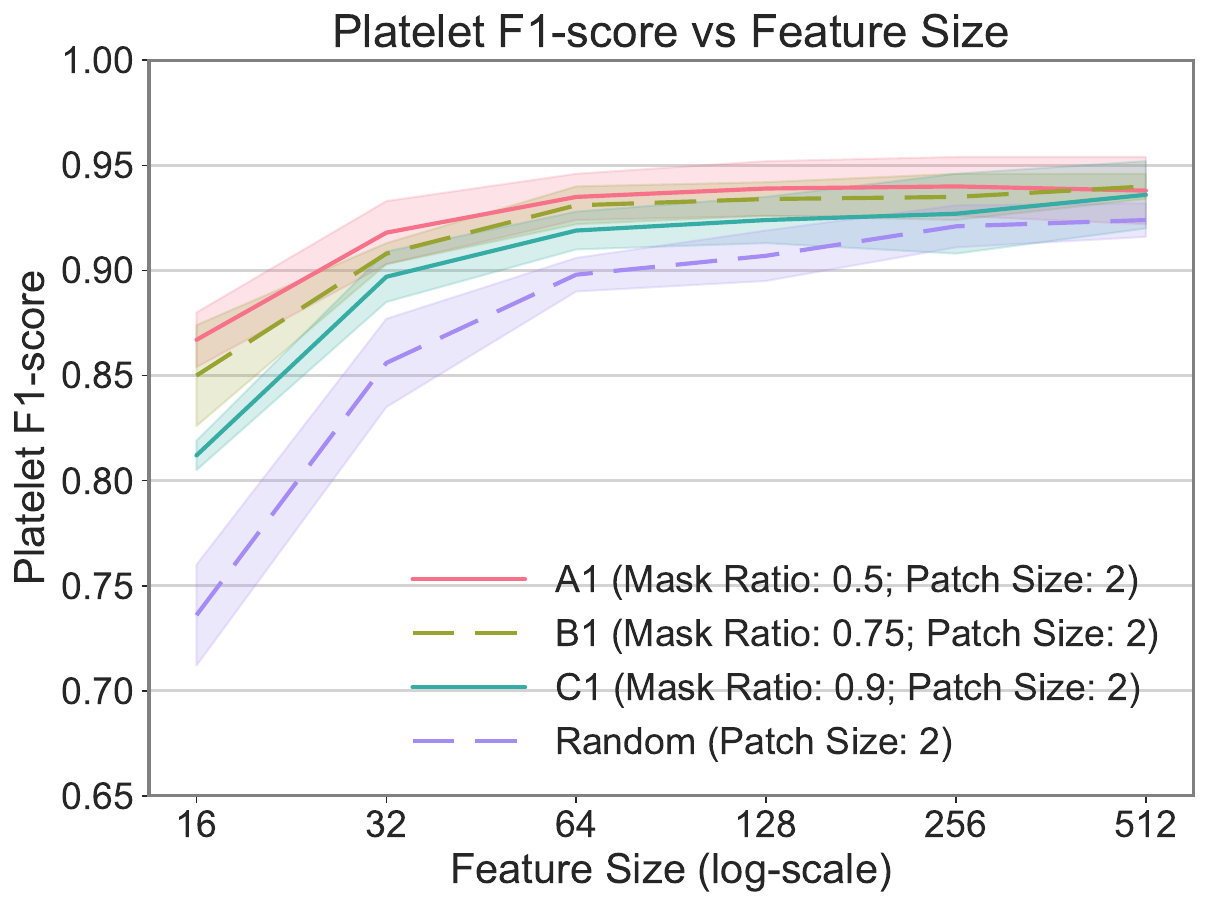}
        \label{fig:f1_platelet_05_75_09_2}
    }
    \hspace{-0.4cm}
    \vspace{-0.2cm}
    \subfigure[]
    {
        \includegraphics[width=0.32\textwidth]{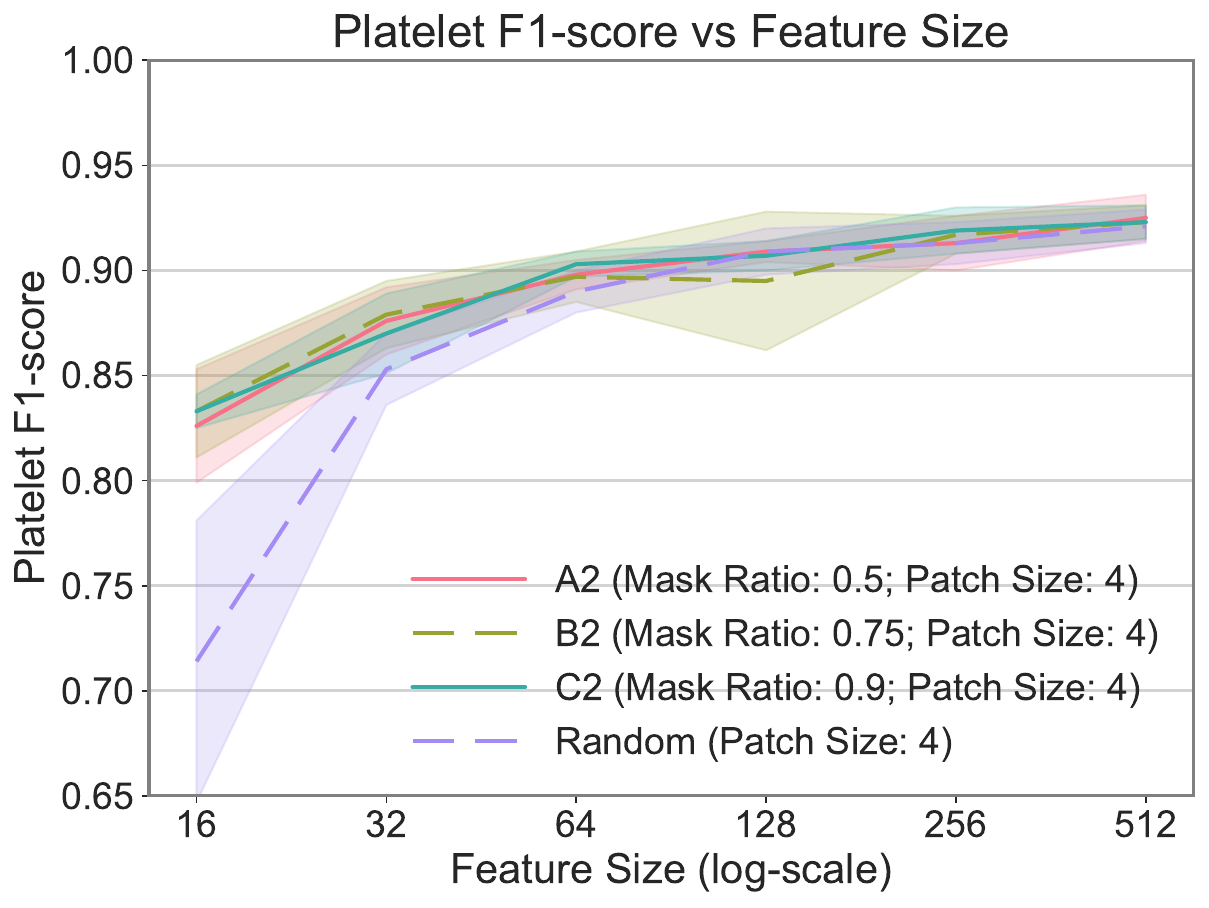}
        \label{fig:f1_platelet_05_75_09_4}
    }
    \hspace{-0.4cm}
    \subfigure[]
    {
        \includegraphics[width=0.32\textwidth]{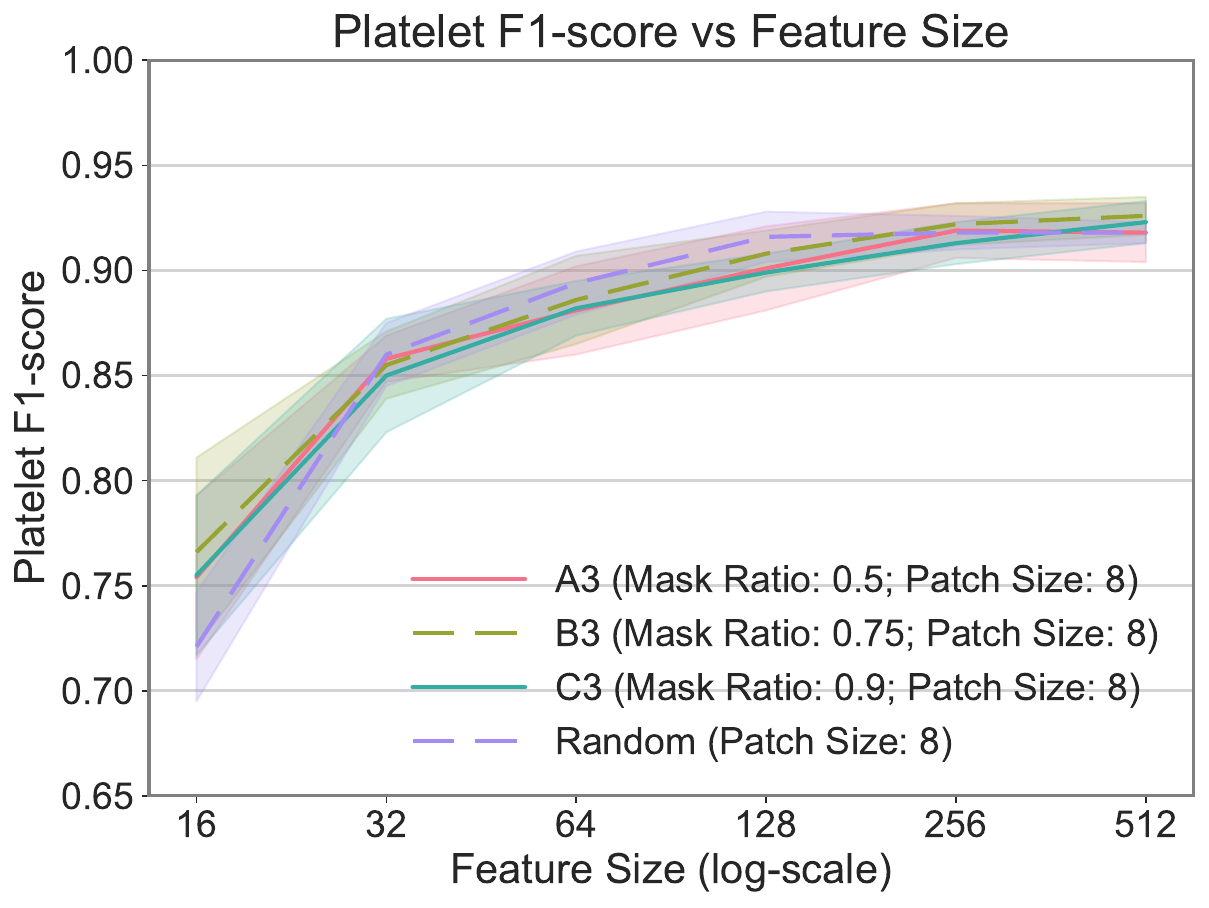}
        \label{fig:f1_platelet_05_75_09_8}
    }
    \caption{Platelet F1-scores for semantic segmentation with different feature sizes using pre-trained encoders (A1--A3, B1--B3, and C1--C3) or randomly initialized encoders (Random: Patch size 2, 4, or 8).
    Shaded areas represent respective standard deviations.}
    \label{fig:f1_platelet_scores}
\end{figure*}

Although a previous report indicated that pre-trained encoders dramatically improved performance~\cite{Zhou::2022a} for image analyses of human lungs, abdomen, and brain tumor tissue samples, our results for blood components suggests otherwise.
In fact, the results from randomly initialized UNETR networks provided similar F1-scores compared to pre-trained networks.
However, for specific blood components such as, platelets, white blood cells, and red blood cells we observed modest improvements of F1-scores for platelets (Fig.~\ref{fig:f1_platelet_scores}).

\begin{table*}[!ht]
\begin{adjustbox}{width=0.98\textwidth}
    \centering
    \begin{threeparttable}
        \caption{Downstream application for semantic segmentation using pre-trained ViT encoders.\tnote{1,2}}
        \begin{tabular}{ l l l l l l l l l }
            \toprule
            % headings
            \textbf{Encoder}
            & \textbf{Feature Size}
            & \textbf{Accuracy}
            & \textbf{F1-score}
            & \multicolumn{5}{c}{\textbf{F1-score}} \\
            \cmidrule(lr){5-9}
            & 
            & 
            & 
            & \textbf{Platelet} 
            & \textbf{WBC} 
            & \textbf{RBC (interior)} 
            & \textbf{RBC (exterior)} 
            & \textbf{Bead}
            \\
            \midrule
            \multicolumn{9}{l}{\textbf{Patch Size: 2}} \\
            \midrule
            % entry: logs/
            A1 (0.5)
            & 512
            & $0.989 \pm 0.002$
            & $0.962 \pm 0.008$
            & $0.938 \pm 0.016$
            & $0.968 \pm 0.011$
            & $0.949 \pm 0.013$
            & $0.981 \pm 0.003$
            & $ 0.981 \pm 0.016$
            \\
            % entry: logs/
            A1 (0.5)
            & 256
            & $0.984 \pm 0.005$
            & $0.959 \pm 0.007$
            & $0.940 \pm 0.014$
            & $0.971 \pm 0.008$
            & $0.947 \pm 0.013$
            & $0.981 \pm 0.004$
            & $0.983 \pm 0.012$
            \\
            % entry: logs/
            A1 (0.5)
            & 128
            & $0.984 \pm 0.005$
            & $0.957 \pm 0.013$
            & $0.939 \pm 0.013$
            & $0.970 \pm 0.009$
            & $0.942 \pm 0.013$
            & $0.978 \pm 0.003$
            & $0.980 \pm 0.007$
            \\
            % entry: logs/
            A1 (0.5)
            & 64
            & $0.983 \pm 0.006$ 
            & $0.953 \pm 0.007$ 
            & $0.935 \pm 0.011$
            & $0.967 \pm 0.010$
            & $0.936 \pm 0.017$
            & $0.974 \pm 0.004$
            & $0.969 \pm 0.015$
            \\
            % entry: logs/
            A1 (0.5)
            & 32
            & $0.982 \pm 0.007$
            & $0.945 \pm 0.005$
            & $0.918 \pm0.015$
            & $0.964 \pm 0.012$
            & $0.928 \pm 0.019$
            & $0.973 \pm 0.007$
            & $0.954 \pm 0.023$
            \\
            % entry: logs/
            A1 (0.5)
            & 16
            & $0.977 \pm 0.004$
            & $0.882 \pm 0.042$
            & $0.867 \pm 0.013$
            & $0.955 \pm 0.015$
            & $0.897 \pm 0.021$
            & $0.962 \pm 0.005$
            & $0.787 \pm 0.194$
            \\
            \midrule
            % entry: logs/
            B1 (0.75)
            & 512
            & $0.985 \pm 0.004$
            & $0.959 \pm 0.006$
            & $0.940 \pm 0.006$
            & $0.966 \pm 0.017$
            & $0.948 \pm 0.014$
            & $0.981 \pm 0.004$
            & $0.979 \pm 0.010$
            \\
            % entry: logs/
            B1 (0.75)
            & 256
            & $0.984 \pm 0.005$
            & $0.958 \pm 0.007$
            & $0.935 \pm 0.011$
            & $0.966 \pm 0.016$
            & $0.947 \pm 0.016$
            & $0.980 \pm 0.005$
            & $0.979 \pm 0.010$
            \\
            % entry: logs/
            B1 (0.75)
            & 128
            & $0.983 \pm 0.002$
            & $0.955 \pm 0.006$
            & $0.934 \pm 0.008$
            & $0.968 \pm 0.013$
            & $0.941 \pm 0.018$
            & $0.976 \pm 0.008$
            & $0.972 \pm 0.015$
            \\
            % entry: logs/
            B1 (0.75)
            & 64
            & $0.980 \pm 0.006$
            & $0.951 \pm 0.011$
            & $0.931 \pm 0.009$
            & $0.964 \pm 0.018$
            & $0.934 \pm 0.018$
            & $0.977 \pm 0.004$
            & $0.965 \pm 0.021$
            \\
            % entry: logs/
            B1 (0.75)
            & 32
            & $0.979 \pm 0.004$
            & $0.942 \pm 0.007$
            & $0.908 \pm 0.005$
            & $0.961 \pm 0.008$
            & $0.925 \pm 0.019$
            & $0.971 \pm 0.005$
            & $0.945 \pm 0.031$
            \\
            % entry: logs/
            B1 (0.75)
            & 16
            & $0.972 \pm 0.007$
            & $0.878 \pm 0.065$
            & $0.850 \pm 0.024$
            & $0.956 \pm 0.014$
            & $0.898 \pm 0.020$
            & $0.962 \pm 0.005$
            & $0.712 \pm 0.382$
            \\
            \midrule
            % entry: logs/
            C1 (0.9)
            & 512
            & $0.986 \pm 0.002$
            & $0.957 \pm 0.013$
            & $0.936 \pm 0.016$
            & $0.966 \pm 0.015$
            & $0.949 \pm 0.013$
            & $0.980 \pm 0.004$
            & $0.976 \pm 0.012$
            \\
            % entry: logs/
            C1 (0.9)
            & 256
            & $0.984 \pm 0.004$
            & $0.954 \pm 0.014$
            & $0.927 \pm 0.019$
            & $0.964 \pm 0.012$
            & $0.943 \pm 0.012$
            & $0.978 \pm 0.005$
            & $0.973 \pm 0.008$
            \\
            % entry: logs/
            C1 (0.9)
            & 128
            & $0.983 \pm 0.005$
            & $0.952 \pm 0.011$
            & $0.924 \pm 0.011$
            & $0.966 \pm 0.017$
            & $0.942 \pm 0.012$
            & $0.979 \pm 0.003$
            & $0.970 \pm 0.012$
            \\
            % entry: logs/
            C1 (0.9)
            & 64
            & $0.982 \pm 0.005$
            & $0.949 \pm 0.009$
            & $0.919 \pm 0.009$
            & $0.965 \pm 0.016$
            & $0.934 \pm 0.013$
            & $0.972 \pm 0.006$
            & $0.965 \pm 0.015$
            \\
            % entry: logs/
            C1 (0.9)
            & 32
            & $0.981 \pm 0.005$
            & $0.933 \pm 0.016$
            & $0.897 \pm 0.012$
            & $0.961 \pm 0.014$
            & $0.924 \pm 0.010$
            & $0.972 \pm 0.005$
            & $0.946 \pm 0.027$
            \\
            % entry: logs/
            C1 (0.9)
            & 16
            & $0.974 \pm 0.003$
            & $0.842 \pm 0.063$
            & $0.812 \pm 0.007$
            & $0.954 \pm 0.014$
            & $0.893 \pm 0.016$
            & $0.961 \pm 0.006$
            & $0.756 \pm 0.094$
            \\
            \midrule
            \multicolumn{9}{l}{\textbf{Patch Size: 4}} \\
            \midrule
            % entry: logs/
            A2 (0.5)
            & 512
            & $0.983 \pm 0.004$
            & $0.951 \pm 0.017$
            & $0.925 \pm 0.011$
            & $0.968 \pm 0.015$
            & $0.943 \pm 0.017$
            & $0.981 \pm 0.003$
            & $0.979 \pm 0.007$
            \\
            % entry: logs/
            A2 (0.5)
            & 256
            & $0.984 \pm 0.004$
            & $0.949 \pm 0.015$
            & $0.913 \pm 0.013$
            & $0.969 \pm 0.011$
            & $0.944 \pm 0.010$
            & $0.980 \pm 0.002$
            & $0.974 \pm 0.014$
            \\
            % entry: logs/
            A2 (0.5)
            & 128
            & $0.984 \pm 0.003$
            & $0.944 \pm 0.020$
            & $0.909 \pm 0.005$
            & $0.971 \pm 0.008$
            & $0.942 \pm 0.011$
            & $0.979 \pm 0.002$
            & $0.971 \pm 0.023$
            \\
            % entry: logs/
            A2 (0.5)
            & 64
            & $0.981 \pm 0.005$
            & $0.937 \pm 0.015$
            & $0.898 \pm 0.007$
            & $0.966 \pm 0.011$
            & $0.929 \pm 0.013$
            & $0.972 \pm 0.004$
            & $0.953 \pm 0.023$
            \\
            % entry: logs/
            A2 (0.5)
            & 32
            & $0.980 \pm 0.005$
            & $0.927 \pm 0.014$
            & $0.876 \pm 0.016$
            & $0.963 \pm 0.012$
            & $0.921 \pm 0.014$
            & $0.972 \pm 0.004$
            & $0.947 \pm 0.015$
            \\
            % entry: logs/
            A2 (0.5)
            & 16
            & $0.977 \pm 0.004$
            & $0.907 \pm 0.013$
            & $0.826 \pm 0.027$
            & $0.957 \pm 0.012$
            & $0.907 \pm 0.022$
            & $0.964 \pm 0.003$
            & $0.889 \pm 0.023$
            \\
            \midrule
            % entry: logs/
            B2 (0.75)
            & 512
            & $0.985 \pm 0.003$
            & $0.953 \pm 0.013$
            & $0.923 \pm 0.008$
            & $0.972 \pm 0.006$
            & $0.940 \pm 0.011$
            & $0.979 \pm 0.002$
            & $0.975 \pm 0.009$
            \\
            % entry: logs/
            B2 (0.75)
            & 256
            & $0.983 \pm 0.005$
            & $0.947 \pm 0.016$
            & $0.917 \pm 0.009$
            & $0.970 \pm 0.011$
            & $0.946 \pm 0.013$
            & $0.979 \pm 0.004$
            & $0.972 \pm 0.011$
            \\
            % entry: logs/
            B2 (0.75)
            & 128
            & $0.983 \pm 0.005$
            & $0.943 \pm 0.018$
            & $0.895 \pm 0.033$
            & $0.969 \pm 0.011$
            & $0.938 \pm 0.010$
            & $0.977 \pm 0.004$
            & $0.972 \pm 0.010$
            \\
            % entry: logs/
            B2 (0.75)
            & 64
            & $0.981 \pm 0.007$
            & $0.940 \pm 0.015$
            & $0.897 \pm 0.012$
            & $0.969 \pm 0.009$
            & $0.929 \pm 0.007$
            & $0.973 \pm 0.005$
            & $0.970 \pm 0.012$
            \\
            % entry: logs/
            B2 (0.75)
            & 32
            & $0.980 \pm 0.006$
            & $0.927 \pm 0.012$
            & $0.879 \pm 0.016$
            & $0.963 \pm 0.017$
            & $0.921 \pm 0.020$
            & $0.968 \pm 0.004$
            & $0.935 \pm 0.017$
            \\
            % entry: logs/
            B2 (0.75)
            & 16
            & $0.976 \pm 0.004$
            & $0.904 \pm 0.018$
            & $0.833 \pm 0.022$
            & $0.952 \pm 0.012$
            & $0.905 \pm 0.016$
            & $0.963 \pm 0.006$
            & $0.894 \pm 0.025$
            \\
            \midrule
            % entry: logs/
            C2 (0.9)
            & 512
            & $0.984 \pm 0.004$
            & $0.950 \pm 0.017$
            & $0.923 \pm 0.008$
            & $0.965 \pm 0.016$
            & $0.944 \pm 0.015$
            & $0.980 \pm 0.002$
            & $0.976 \pm 0.008$
            \\
            % entry: logs/
            C2 (0.9)
            & 256
            & $0.984 \pm 0.004$
            & $0.948 \pm 0.017$
            & $0.919 \pm 0.011$
            & $0.967 \pm 0.014$
            & $0.942 \pm 0.012$
            & $0.977 \pm 0.004$
            & $0.971 \pm 0.011$
            \\
            % entry: logs/
            C2 (0.9)
            & 128
            & $0.984 \pm 0.005$
            & $0.947 \pm 0.016$
            & $0.907 \pm 0.007$
            & $0.965 \pm 0.016$
            & $0.943 \pm 0.010$
            & $0.978 \pm 0.003$
            & $0.965 \pm 0.016$
            \\
            % entry: logs/
            C2 (0.9)
            & 64
            & $0.983 \pm 0.005$
            & $0.939 \pm 0.018$
            & $0.903 \pm 0.006$
            & $0.961 \pm 0.014$
            & $0.931 \pm 0.013$
            & $0.974 \pm 0.002$
            & $0.966 \pm 0.014$
            \\
            % entry: logs/
            C2 (0.9)
            & 32
            & $0.980 \pm 0.005$
            & $0.926 \pm 0.007$
            & $0.870 \pm 0.019$
            & $0.959 \pm 0.014$
            & $0.916 \pm 0.015$
            & $0.969 \pm 0.006$
            & $0.946 \pm 0.012$
            \\
            % entry: logs/
            C2 (0.9)
            & 16
            & $0.977 \pm 0.003$
            & $0.904 \pm 0.022$
            & $0.833 \pm 0.008$
            & $0.949 \pm 0.015$
            & $0.904 \pm 0.019$
            & $0.962 \pm 0.003$
            & $0.882 \pm 0.019$
            \\
            \midrule
            \multicolumn{9}{l}{\textbf{Patch Size: 8}} \\
            \midrule
            % entry: logs/
            A3 (0.5)
            & 512
            & $0.981 \pm 0.004$
            & $0.947 \pm 0.016$
            & $0.918 \pm 0.014$
            & $0.959 \pm 0.015$
            & $0.940 \pm 0.009$
            & $0.974 \pm 0.011$
            & $0.979 \pm 0.013$
            \\
            % entry: logs/
            A3 (0.5)
            & 256
            & $0.981 \pm 0.006$
            & $0.945 \pm 0.020$
            & $0.919 \pm 0.013$
            & $0.962 \pm 0.011$
            & $0.941 \pm 0.008$
            & $0.979 \pm 0.003$
            & $0.975 \pm 0.006$
            \\
            % entry: logs/
            A3 (0.5)
            & 128
            & $0.980 \pm 0.007$
            & $0.942 \pm 0.018$
            & $0.901 \pm 0.020$
            & $0.958 \pm 0.011$
            & $0.936 \pm 0.007$
            & $0.977 \pm 0.004$
            & $0.969 \pm 0.011$
            \\
            % entry: logs/
            A3 (0.5)
            & 64
            & $0.977 \pm 0.005$
            & $0.934 \pm 0.013$
            & $0.881 \pm 0.021$
            & $0.949 \pm 0.015$
            & $0.927 \pm 0.008$
            & $0.971 \pm 0.002$
            & $0.955 \pm 0.019$
            \\
            % entry: logs/
            A3 (0.5)
            & 32
            & $0.973 \pm 0.008$
            & $0.916 \pm 0.015$
            & $0.858 \pm 0.011$
            & $0.942 \pm 0.021$
            & $0.907 \pm 0.019$
            & $0.963 \pm 0.009$
            & $0.932 \pm 0.035$
            \\
            % entry: logs/
            A3 (0.5)
            & 16
            & $0.963 \pm 0.008$
            & $0.866 \pm 0.013$
            & $0.754 \pm 0.039$
            & $0.925 \pm 0.015$
            & $0.878 \pm 0.026$
            & $0.949 \pm 0.002$
            & $0.850 \pm 0.052$
            \\
            \midrule
            % entry: logs/
            B3 (0.75)
            & 512
            & $0.981 \pm 0.006$
            & $0.946 \pm 0.018$
            & $0.926 \pm 0.009$
            & $0.958 \pm 0.015$
            & $0.944 \pm 0.016$
            & $0.978 \pm 0.006$
            & $0.976 \pm 0.010$
            \\
            % entry: logs/
            B3 (0.75)
            & 256
            & $0.983 \pm 0.004$
            & $0.946 \pm 0.017$
            & $0.922 \pm 0.010$
            & $0.960 \pm 0.015$
            & $0.940 \pm 0.010$
            & $0.979 \pm 0.004$
            & $0.973 \pm 0.007$
            \\
            % entry: logs/
            B3 (0.75)
            & 128
            & $0.981 \pm 0.005$
            & $0.942 \pm 0.017$
            & $0.908 \pm 0.011$
            & $0.961 \pm 0.013$
            & $0.934 \pm 0.009$
            & $0.978 \pm 0.002$
            & $0.965 \pm 0.012$
            \\
            % entry: logs/
            B3 (0.75)
            & 64
            & $0.978 \pm 0.004$
            & $0.933 \pm 0.013$
            & $0.886 \pm 0.021$
            & $0.957 \pm 0.011$
            & $0.923 \pm 0.019$
            & $0.970 \pm 0.004$
            & $0.958 \pm 0.015$
            \\
            % entry: logs/
            B3 (0.75)
            & 32
            & $0.972 \pm 0.005$
            & $0.918 \pm 0.018$
            & $0.855 \pm 0.016$
            & $0.943 \pm 0.012$
            & $0.914 \pm 0.013$
            & $0.964 \pm 0.009$
            & $0.922 \pm 0.026$
            \\
            % entry: logs/
            B3 (0.75)
            & 16
            & $0.956 \pm 0.008$
            & $0.866 \pm 0.014$
            & $0.766 \pm 0.045$
            & $0.922 \pm 0.021$
            & $0.859 \pm 0.044$
            & $0.945 \pm 0.013$
            & $0.869 \pm 0.027$
            \\
            \midrule
            % entry: logs/
            C3 (0.9)
            & 512
            & $0.981 \pm 0.005$
            & $0.945 \pm 0.015$
            & $0.923 \pm 0.010$
            & $0.964 \pm 0.012$
            & $0.943 \pm 0.015$
            & $0.976 \pm 0.009$
            & $0.976 \pm 0.006$
            \\
            % entry: logs/
            C3 (0.9)
            & 256
            & $0.982 \pm 0.004$
            & $0.944 \pm 0.011$
            & $0.913 \pm 0.010$
            & $0.961 \pm 0.016$
            & $0.940 \pm 0.014$
            & $0.979 \pm 0.003$
            & $0.973 \pm 0.014$
            \\
            % entry: logs/
            C3 (0.9)
            & 128
            & $0.978 \pm 0.006$
            & $0.941 \pm 0.019$
            & $0.899 \pm 0.009$
            & $0.959 \pm 0.013$
            & $0.937 \pm 0.006$
            & $0.976 \pm 0.006$
            & $0.971 \pm 0.011$
            \\
            % entry: logs/
            C3 (0.9)
            & 64
            & $0.975 \pm 0.008$
            & $0.927 \pm 0.025$
            & $0.882 \pm 0.013$
            & $0.955 \pm 0.016$
            & $0.920 \pm 0.016$
            & $0.971 \pm 0.004$
            & $0.952 \pm 0.013$
            \\
            % entry: logs/
            C3 (0.9)
            & 32
            & $0.966 \pm 0.008$
            & $0.907 \pm 0.026$
            & $0.850 \pm 0.027$
            & $0.941 \pm 0.017$
            & $0.912 \pm 0.015$
            & $0.964 \pm 0.004$
            & $0.920 \pm 0.028$
            \\
            % entry: logs/
            C3 (0.9)
            & 16
            & $0.957 \pm 0.013$
            & $0.854 \pm 0.022$
            & $0.755 \pm 0.038$
            & $0.929 \pm 0.014$
            & $0.847 \pm 0.052$
            & $0.941 \pm 0.011$
            & $0.841 \pm 0.052$
            \\
            \midrule
            \bottomrule
        \end{tabular}
        $^1$ Models used learned encoder weights from corresponding ViT-MAE networks (mask ratios in parentheses) by loading weights into the encoder-component of a UNETR network.
        Training was done over 200 epochs with the encoder frozen for the first 100 epochs. 
        $^2$ Metrics represents 5-fold cross-validation providing mean values along with their corresponding standard deviations.
        Values for accuracy and F1-scores represents all 9 classes.
        Individual background, artifacts, debris, and bubbles omitted for clarity.
        \label{tab:unetr_combined}
    \end{threeparttable}
\end{adjustbox}
\end{table*}

% \FloatBarrier

% \FloatBarrier

\begin{figure*}[!htb]
    \centering
    \subfigure
    {
        \includegraphics[width=0.32\textwidth]{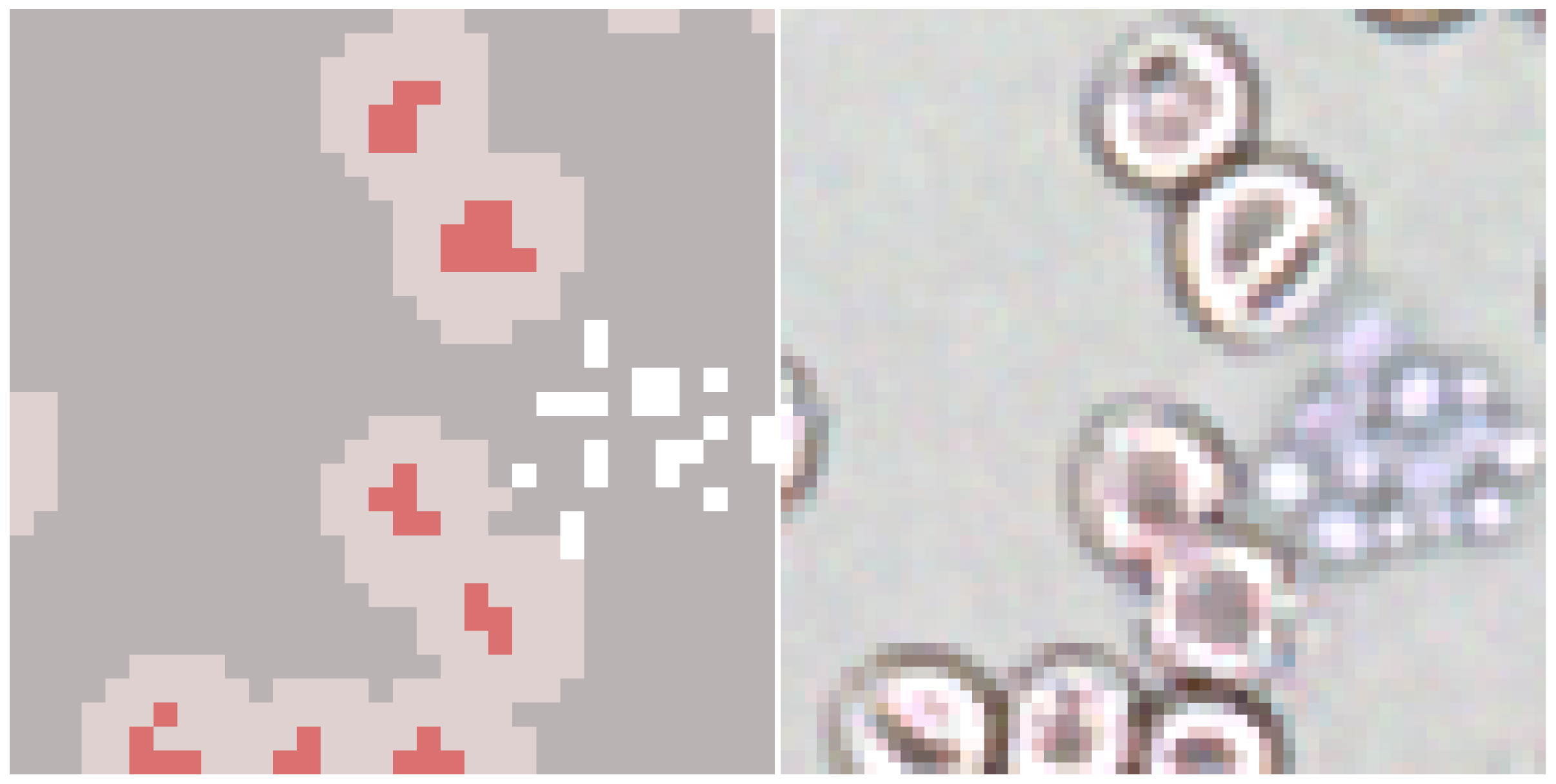}
        \label{fig:unetr_random_segment_02_2_2}
    }
    \vspace{-0.2cm}
    \hspace{-0.4cm}
    \subfigure
    {
        \includegraphics[width=0.32\textwidth]{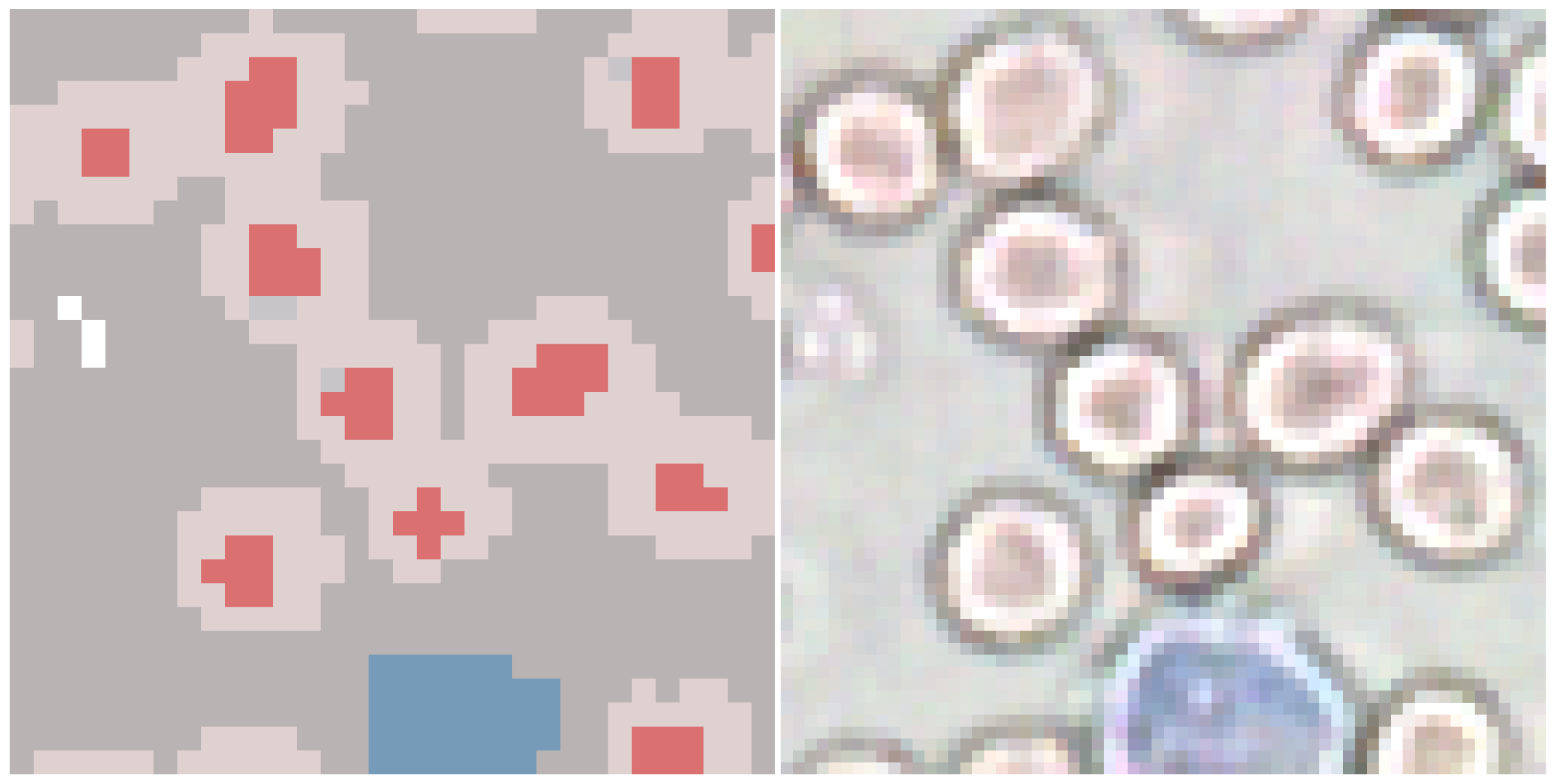}
        \label{fig:unetr_random_segment_02_2_389}
    }
    \hspace{-0.4cm}
    \subfigure
    {
        \includegraphics[width=0.32\textwidth]{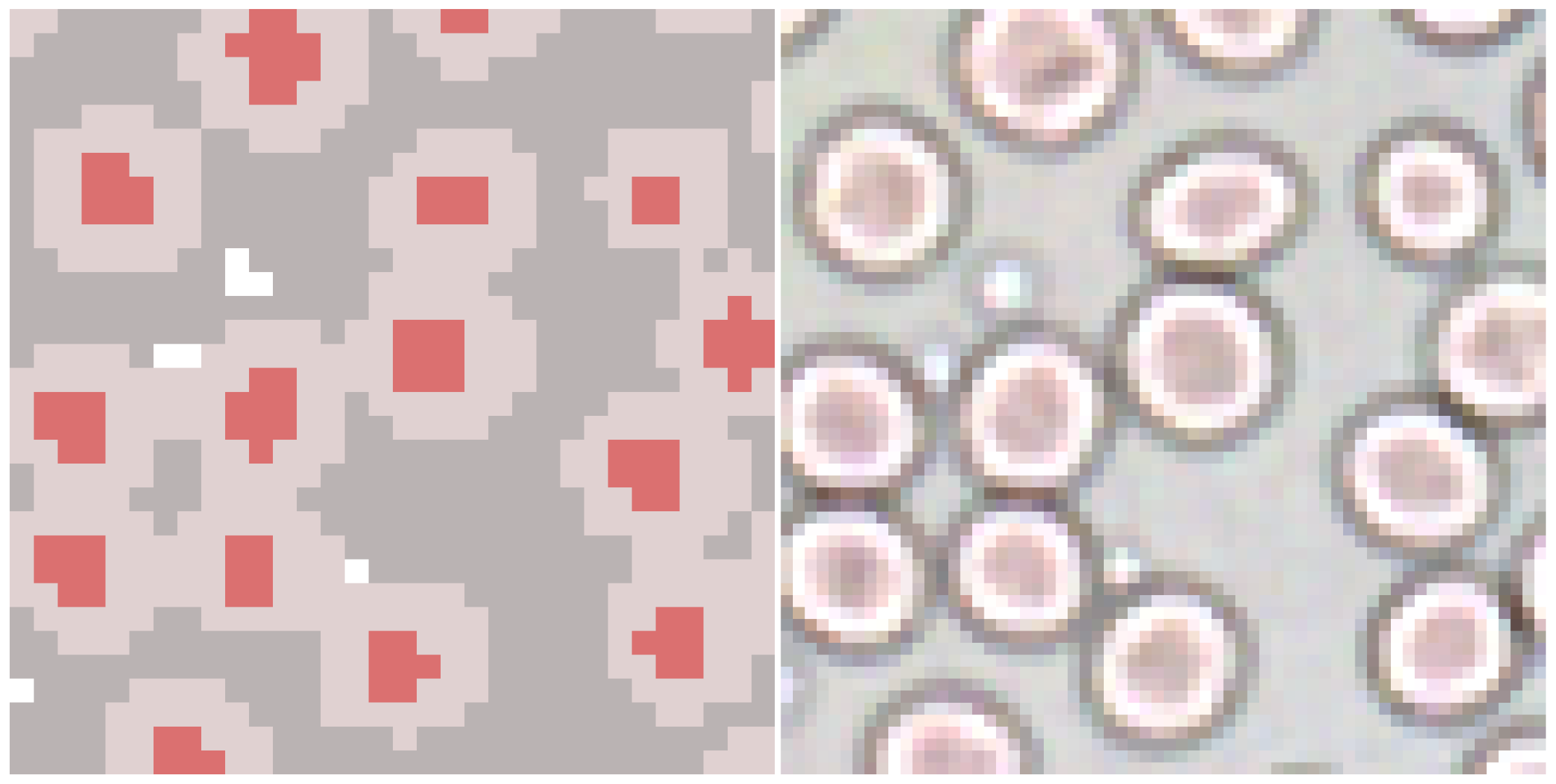}
        \label{fig:unetr_random_segment_02_2_405}
    }
    \subfigure
    {
        \includegraphics[width=0.32\textwidth]{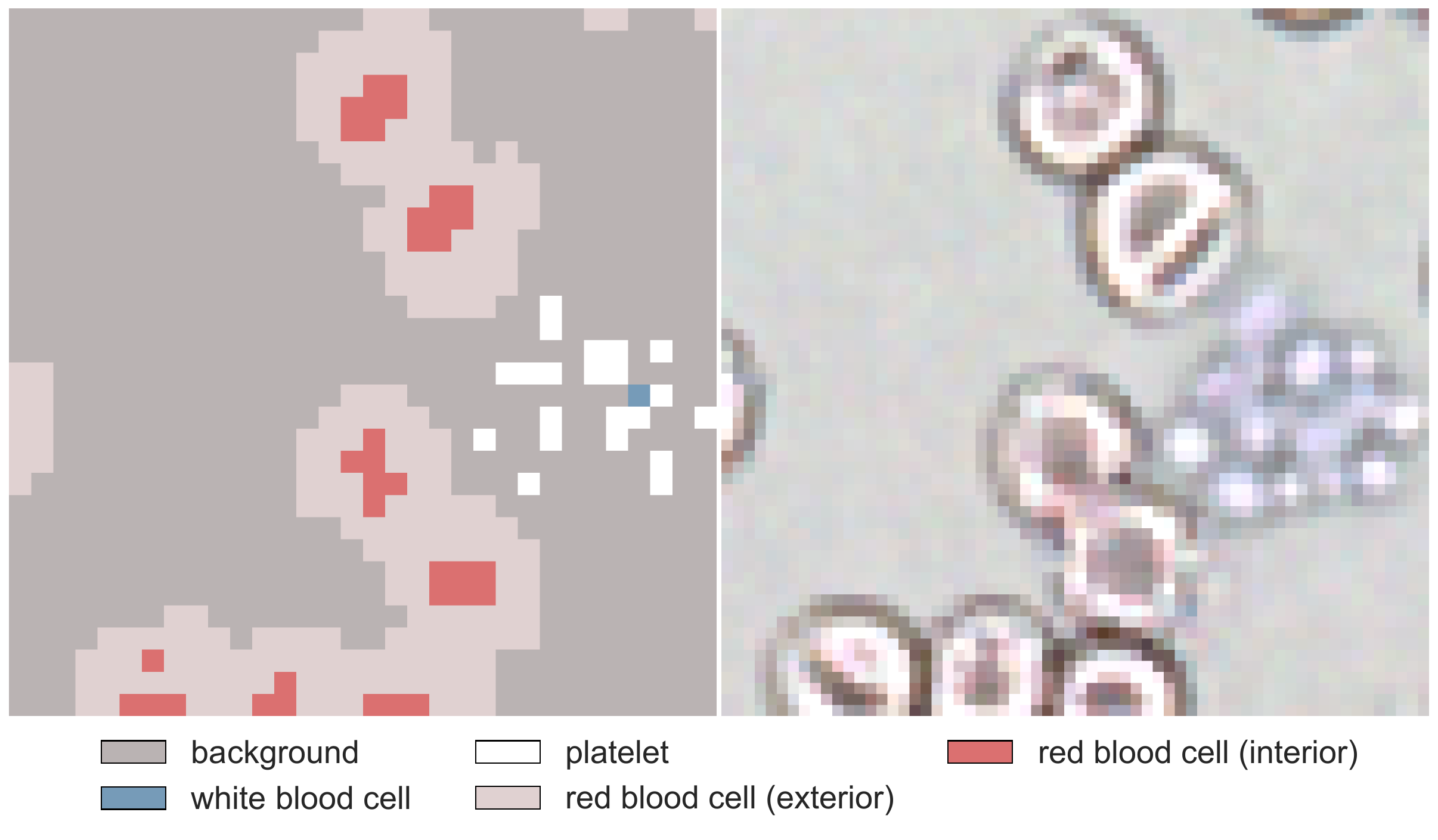}
        \label{fig:unetr_segment_02_2_2}
    }
    \hspace{-0.4cm}
    \subfigure
    {
        \includegraphics[width=0.32\textwidth]{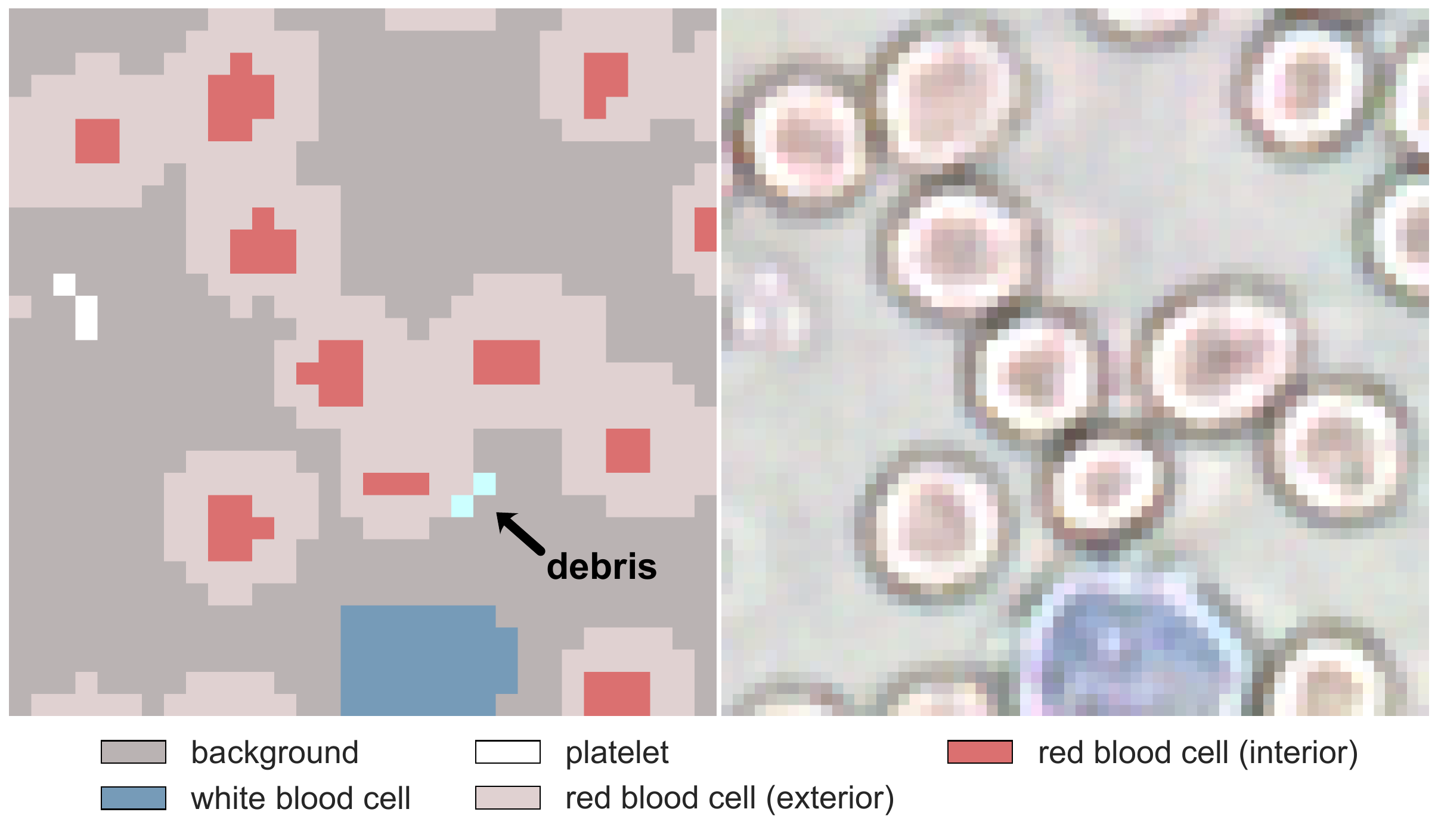}
        \label{fig:unetr_segment_02_2_389}
    }
    \hspace{-0.4cm}
    \subfigure
    {
        \includegraphics[width=0.32\textwidth]{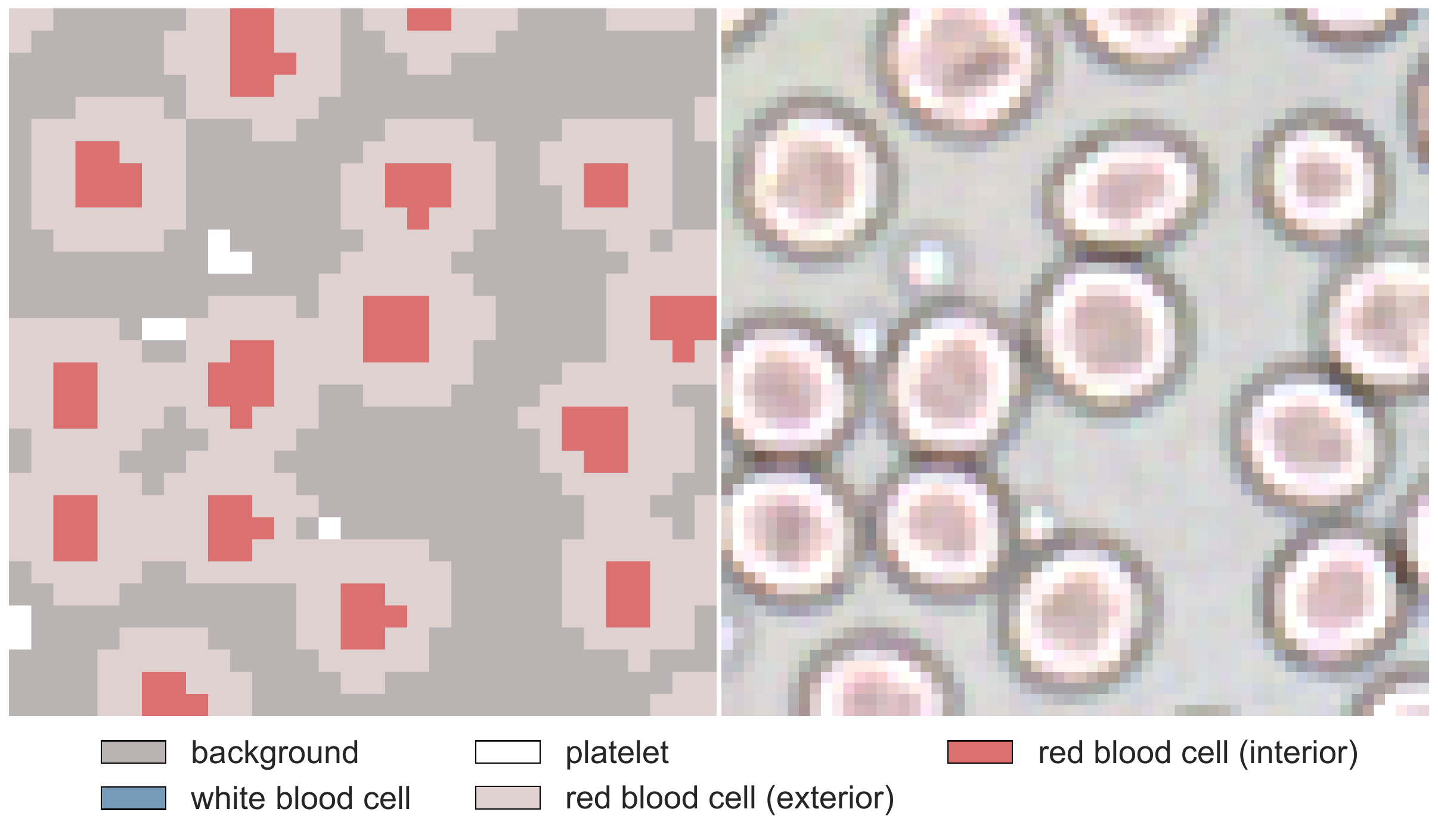}
        \label{fig:unetr_segment_02_2_405}
    }
    \vspace{-0.2cm}
    \hspace{-0.1cm}
    \caption{Segmentation using pre-trained (first row) or randomly initialized (second row) encoders for UNETR networks using a feature size of 64.
    Each row consists of three sets of center-cropped ($32 \times 32$) segmentation examples and corresponding super-resolved original images ($128 \times 128 \times 3$).
    Row 1: [pre-trained encoder] mask ratio 0.5, patch size 2, feature size 64.
    Row 2: [randomly initialized encoder] patch size 2, feature size 64.}
    \label{fig:segmentation_results}
\end{figure*}

\FloatBarrier

\noindent
As mentioned previously in Section~\ref{sect:related}, platelets are difficult to classify.
Figure~\ref{fig:f1_platelet_scores} shows F1-scores for varying feature size for different encoders used for semantic segmentation.
In the first panel (Fig.~\ref{fig:f1_platelet_05_75_09_2}), encoders A1, B1, and C1 (patch size 2) provided similar results; however, these results are statistically significant with respect to randomly initialized encoders when using features size up to 16, 32, and 64.
This suggests that pre-trained encoders with these configurations does improve performance.
For patch sizes 4 and 8 (Fig.~\ref{fig:f1_platelet_05_75_09_4} and Fig.~\ref{fig:f1_platelet_05_75_09_8}), this observation was only present when using feature size 16 and patch size 4 (Fig.~\ref{fig:f1_platelet_05_75_09_4}). 
Overall, these results indicate that pre-training is beneficial for the semantic segmentation of platelets where encoders are trained using a small mask ratio and patch size.
Furthermore, there is no advantage to having large feature sizes to capture additional local spatial context for the UNETR networks.
Moving from a network using 64 (3.1M trainable parameters) to 128 (3.7M trainable parameters) feature size did not produce any significant improvements.
Suggesting that networks with feature sizes greater than 64 may not be necessary.
Comparing the F1-scores by increasing the patch size (Fig.~\ref{fig:f1_platelet_05_75_09_2} vs.~Fig.~\ref{fig:f1_platelet_05_75_09_4} vs.~Fig.~\ref{fig:f1_platelet_05_75_09_8}), we observed that F1-scores are converging to the results obtained using a randomly initialized UNETR network.
Recall that platelet sizes can be as small as a single pixel, exist as aggregates, and features difficult to distinguish from nearby blood components.
When using large patch sizes, networks are challenged to capture and maintain information for platelets. 
For components with well-defined and consistent characteristics (\textit{e.g.}, beads), UNETR networks able maintain representations for semantic segmentation from spatial contexts used with skip connections and upsampled representation(s) from the encoder.

\subsection{Comparison of Segmentation Masks.}

In this section, we provide three examples of segmentation masks and ground truth images for pre-trained and randomly initialized networks with a patch size 2 and encoder derived from a mask ratio of 0.5 (Fig.~\ref{fig:segmentation_results}).
The first column compares an example with a large platelet aggregate.
Both networks provided very similar results; however, the random network predicted a single pixel as a white blood cell.
For medical applications, this minor improvement has the benefit of increasing the accuracy of complete blood counts.
For smaller platelet aggregates (second column), both pre-trained and randomly initialized networks provided the same result.
In this case, the randomly initialized network also picked of debris (black arrow) which was not present for the pre-trained network.
This suggest there is an additional benefits for pre-trained networks to resolve non-cellular structure.
For individual platelets (last column), both networks provide the same predictions.

% CONCLUSION
\section{Conclusions}
\label{sect:conclusions}
This study introduced the use of lensless near-field microscopy to obtain high-resolution images with a large field of view for blood component analyses.
We trained UNETR networks for baseline semantic segmentation of blood components that provided F1-scores up to 95\%.
After training several MAE networks, using different mask ratios and patch sizes, when extracted encoder weights and applied pre-training of UNETR networks for classification.
For overall F1-scores, we observed no statistically significant results from using pre-trained encoder.
However, in the case of small blood components (\textit{e.g.}, platelets), we observed improved results using patch size 2 and mask ratio of 0.5 for pre-training encoders.
While we discussed small objects in the context of blood components, we expect that these conclusions carry over to any small objects in large images.

% REFERENCES
%% The file IEEEtran.bst is a bibliography style file for BibTeX 0.99c (IJCNN)

\bibliographystyle{IEEEtran}
\bibliography{references}

\end{document}